\documentclass[runningheads]{llncs}
\usepackage{fontspec}
\usepackage{graphicx}
\usepackage{booktabs}
\usepackage{amsmath,amssymb}
\usepackage{algorithm}
\usepackage{algpseudocode}
\usepackage[table]{xcolor}
\definecolor{Forestgreen}{rgb}{0.13,0.55,0.13}
\usepackage{multirow}
\usepackage{enumitem}
\usepackage{arydshln}
\usepackage{hyperref}
\usepackage{adjustbox}
\usepackage[utf8]{inputenc}
\usepackage{fontenc}
\usepackage{polyglossia}
\usepackage{placeins}
\setmainlanguage{english}
\setotherlanguage{hindi}
\newfontfamily\devanagarifont{Eczar-Regular.otf}[Script=Devanagari]
\newcommand{\charbm}{\textsc{CharBM25}}

\usepackage{tcolorbox}
\tcbuselibrary{breakable,skins}
\tcbset{
  promptbox/.style={
    breakable,
    colback=gray!6,
    colframe=gray!45,
    boxrule=0.4pt,
    arc=2pt,
    left=6pt, right=6pt, top=4pt, bottom=4pt,
    fontupper=\small\ttfamily,
    title style={colback=gray!20},
  }
}

\begin{document}

\title{Evaluating In-Context Learning and Retrieval Strategies for Devanagari Post-OCR Correction}

\titlerunning{CharBM25 for Devanagari Post-OCR Correction}

\author{Abhishek Bhandari\orcidID{0000-0001-5193-7213}\and Gaurav Harit\orcidID{0000-0001-7943-0123}}
\authorrunning{Bhandari and Harit}
\institute{Indian Institute of Technology Jodhpur, India\\
\email{bhandari.1@ iitj.ac.in, gharit@iitj.ac.in}}

\maketitle

\begin{abstract}
In-context learning using Large Language Models (LLMs) offers a compelling path to training-free post-OCR correction, yet its effectiveness for Devanagari script remains entirely unexplored. We present the first systematic evaluation of LLMs (3B–32B) for post-OCR correction in Hindi and Marathi, comparing three in-context example retrieval strategies: domain-random selection, dense semantic retrieval, and our proposed \charbm{}. \charbm{} retrieves examples by character n-gram BM25 similarity over OCR inputs, targeting shared error patterns between retrieved examples and the test sentence. Across a 20,000-sentence benchmark spanning five news domains, retrieval strategy is the decisive factor in correction quality: \charbm{} outperforms domain-random selection by 2.8--4.0\,pp absolute WER on Hindi (2.9--3.8\,pp on Marathi). Scale dominates performance: Gemma-3-27B achieves WER reductions of 55.0\% for Hindi and 33.3\% for Marathi under \charbm{}-5 (n=3). Few-shot gains are capacity-gated: models below 8B do not reliably improve over the OCR baseline, and on Marathi, the smallest models (3B) degrade more sentences than they improve. Models of 12B or more achieve Hindi improvement rates of 76–94\%, though Marathi rates vary more widely (42–78\%), reflecting its greater morphological complexity. Marathi is persistently harder to correct: the gap in WER reduction between the two languages grows with scale, from 8.8\% at 3B (\charbm{}-5: $\Delta${WER} 9.2\% Hindi vs.\ 0.4\% Marathi) to 21.7\% at 27B (55.0\% vs.\ 33.3\%). Trigrams consistently outperform bigrams and unigrams across all models and both languages. These findings establish \charbm{} as an effective, GPU-free retrieval strategy that outperforms random selection and matches or exceeds dense retrieval at negligible computational cost. Combining \charbm{} with a general-purpose LLM of 12B parameters or more delivers reliable, training-free Devanagari post-OCR correction without task-specific fine-tuning. The dataset is made publicly available at \url{https://huggingface.co/datasets/AbhishekBhandari/Devanagari-OCR-ICL-Benchmark}.
\end{abstract}

\section{Introduction}
\label{sec:intro}

Digitizing printed Devanagari documents is essential for preserving cultural heritage and enabling downstream NLP for Hindi and Marathi, two of the most widely spoken Indic languages. OCR engines produce
high error rates on Devanagari due to three script-specific challenges. First, dependent vowel marks (\emph{matras}) attach to consonants, and changing a single mark alters a word entirely --- for example, \texthindi{कि} (ki) and \texthindi{की} (k\=\i) differ by only one mark yet are distinct words (Table~\ref{tab:char_pairs}). Second,
conjunct consonants are written as fused ligatures, and OCR frequently fragments them into their component consonants, producing output that is orthographically plausible but semantically wrong (Table~\ref{tab:conjunct}). Third, many Devanagari characters are visually near-identical and are confused systematically by the recognizer (Table~\ref{tab:char_pairs}).

Post-OCR correction has been addressed primarily using fine-tuned seq2seq models~\cite{kashid2025roundtripocr,10.1145/3815575}, which require task-specific training data and considerable compute resources. In-context learning (ICL) is a prompting paradigm in which a language model is presented with a small number of (input, output) demonstration pairs — called shots — directly in the prompt, and is expected to infer the task from these examples and generalize to new inputs without updating any model parameters. ICL thus offers an attractive alternative to fine-tuning for post-OCR correction: given a few (OCR-corrupted sentence, corrected sentence) pairs as demonstrations, a prompted LLM may correct new OCR inputs with no task-specific training at all. English-language studies show both promise~\cite{thomas2024leveraging} and limitations~\cite{kanerva2025ocr,boros2024postcorrection} for this approach, but whether it transfers to Devanagari remains entirely open.
\begin{table}[t]
\centering
\small
\begin{tabular}{@{}ccc@{\hspace{1.5em}}ccc@{}}
\toprule
\multicolumn{3}{c}{\textbf{Consonant pairs}} &
\multicolumn{3}{c}{\textbf{Matra / diacritic pairs}} \\
\cmidrule(lr){1-3}\cmidrule(lr){4-6}
\texthindi{ब} $\leftrightarrow$ \texthindi{व} & ba/va & U+092C/0935 &
\texthindi{कि} $\leftrightarrow$ \texthindi{की} & ki/k\=\i & U+093F/0940 \\
\texthindi{म} $\leftrightarrow$ \texthindi{भ} & ma/bha & U+092E/092D &
\texthindi{कु} $\leftrightarrow$ \texthindi{कू} & ku/k\=u & U+0941/0942 \\
\texthindi{ण} $\leftrightarrow$ \texthindi{न} & \d{n}a/na & U+0923/0928 &
\texthindi{कं} $\leftrightarrow$ \texthindi{कँ} & anusvara/ & U+0902/0901 \\
\texthindi{ध} $\leftrightarrow$ \texthindi{घ} & dha/gha & U+0927/0918 &
 & chandrabindu & \\
\bottomrule
\end{tabular}
\caption{Visually similar Devanagari character pairs that Tesseract
confuses systematically. Each pair differs by a single stroke or
diacritic that is easily lost at scan resolution.}
\label{tab:char_pairs}
\end{table}

The central design question in ICL is \emph{example selection}: which sentences from the shot bank should serve as demonstrations? For OCR correction, this choice is non-trivial because the corrective value of an example depends on how closely its error structure matches that of the test input. We study three retrieval strategies that differ in what they optimize: domain-random selection, dense semantic retrieval (embedding similarity), and \charbm{} (character n-gram BM25 similarity over OCR inputs). Semantic similarity is a natural and effective retrieval criterion for tasks like translation or summarisation. For OCR correction, however, an additional and more direct signal is available: an OCR engine does not misread a word because of its meaning --- it misreads it because a matra stroke is ambiguous at low resolution, or because a conjunct ligature is split into its component consonants by the segmenter. Two sentences on completely unrelated topics may share the same OCR error pattern (e.g., both contain the confusion 
\texthindi{ब}→\texthindi{व}), so a demonstration that resolves that pattern provides direct corrective guidance regardless of semantic domain. \charbm{} operationalizes this insight: it retrieves examples by character n-gram BM25 similarity over the OCR input side, targeting shared surface-level corruption patterns directly --- and, as our results show, doing so on par with dense retrieval while requiring no embedding model or GPU.

\begin{table}[t]
\centering
\small
\begin{tabular}{@{}llll@{}}
\toprule
\textbf{Conjunct (GT)} & \textbf{OCR output} & \textbf{Codepoint change} & \textbf{Effect on word} \\
\midrule
\texthindi{\textcolor{green}{क्ष}} & \texthindi{\textcolor{red}{क ष}}
  & virama (U+094D) dropped
  & \texthindi{क्षमा} $\to$ \texthindi{क षमा} \\
\texthindi{\textcolor{green}{प्र}} & \texthindi{\textcolor{red}{प र}}
  & subscript-r split off
  & \texthindi{प्रण} $\to$ \texthindi{प रण} \\
\bottomrule
\end{tabular}
\caption{Conjunct consonant fragmentation: OCR splits a ligated
conjunct into its component consonants, dropping the virama.
The result is orthographically legal but semantically broken.}
\label{tab:conjunct}
\end{table}
\paragraph{Research Questions.}
\begin{enumerate}[label=\textbf{RQ\arabic*:}, leftmargin=*, noitemsep]
  \item How well do LLMs correct Devanagari OCR errors under zero-shot
        prompting, and to what extent does in-context learning improve
        over zero-shot across model scales?
   \item How do the three retrieval strategies --- domain-random selection, dense semantic retrieval, and \charbm{} --- compare, and which $n$-gram order is most effective for \charbm{}?
  \item Is few-shot correction uniformly beneficial, or do some sentences improve while others degrade? Is there a minimum model capacity — measured in parameter count — below which \charbm{} demonstrations fail to produce net improvement across the test set, and models degrade more sentences than they correct?
  \item How do gains differ across Hindi and Marathi, and how does
        the language gap evolve with model scale?
\end{enumerate}

\paragraph{Contributions.}
\begin{enumerate}[label=(\roman*), leftmargin=*, noitemsep]
  \item A \textbf{20{,}000-sentence Devanagari post-OCR benchmark}
        spanning five news domains for Hindi and Marathi, with
        OCR baselines of WER 36.06\% (Hindi) and 35.11\% (Marathi)
        (\S\ref{sec:dataset}).
  \item \textbf{\charbm{}}---a character $n$-gram BM25 retrieval
        algorithm for in-context example selection, with ablation
        over $n \in \{1,2,3\}$ across all 11 models and both
        languages (\S\ref{sec:charbm25}).
  \item A \textbf{systematic evaluation} of eleven open-weight LLMs spanning 3B–32B parameters — Llama-3.2-3B, Qwen3-4B, Gemma-3-4B, Llama-3.1-8B, Qwen3-8B, Aya-101, Gemma-3-12B, Sarvam-m, Qwen3-14B, Gemma-3-27B, and Qwen3-32B — under four retrieval settings: zero-shot (no demonstrations), domain-random selection, \charbm{}, and dense retrieval (cosine similarity over sentence embeddings using L3Cube-HindSBERT for Hindi and L3Cube-MahaSBERT for Marathi), with
        per-sentence benefit analysis and language asymmetry
        quantification (\S\ref{sec:results}).
\end{enumerate}


\section{Related Work}
\label{sec:related}

\subsection{Post-OCR correction for Indic scripts}
Post-OCR correction has been studied across several Indic scripts. For Bangla, Pal and Chaudhuri~\cite{pal2000ocr} applied dictionary-based word-level correction using morphological analysis, and Hasan et al.~\cite{7916243} extended this with a two-stage pipeline combining character-level confusion matrices with a statistical language model, achieving substantial WER reduction on printed Bangla documents. For Gurmukhi, Gill et al.~\cite{953957} proposed a lexicon-driven post-processing stage that corrects character substitutions arising from visually confusable glyph pairs in Gurumukhi script. For Malayalam, Antony and Chaudhuri~\cite{10.1145/1815330.1815394} 
developed a morphology-aware correction module exploiting the agglutinative structure of Malayalam to constrain the candidate correction space. For Devanagari specifically, early methods relied
on dictionary partitioning and character confusion matrices~\cite{bansal2002partitioning}, later extended with frameworks that learn $n$-gram confusion patterns from human feedback~\cite{saluja2017framework} and LSTM-based character language
models~\cite{saluja2017error}.
Neural seq2seq models with copy mechanisms improved upon rule-based
methods for Sanskrit~\cite{krishna2018upcycle}, and Maheshwari et
al.~\cite{maheshwari2022benchmark} established a benchmark and
fine-tuning comparisons across MBART, mT5, ByT5, and IndicBART for
Sanskrit post-OCR correction.
For scalable data creation, Kashid and
Bhattacharyya~\cite{kashid2025roundtripocr} proposed RoundTripOCR, a
render-and-OCR pipeline for Devanagari that generates training pairs
across six languages without manual annotation.
Guan and Greene~\cite{guan-greene-2024-advancing} demonstrated that glyph-similarity-based synthetic data substantially reduces ByT5 error
rates across multiple scripts, including Telugu. Bhandari and
Harit~\cite{10.1145/3815575}, showed that providing the corrected
preceding sentence as an auxiliary encoder input to an MBART-50
model reduces Hindi WER from 28.53\% to 14.57\% and Marathi WER from
24.89\% to 8.28\%. Our work asks a complementary question: whether \emph{prompted} LLMs,
without any fine-tuning, can approach the WER reductions achieved
through better few-shot example selection.

\subsection{LLMs for post-OCR correction}
Thomas et al.~\cite{thomas2024leveraging} showed that fine-tuning Llama~2
on historical English newspapers reduces CER by 54\%, substantially
outperforming BART.
Kanerva et al.~\cite{kanerva2025ocr} evaluated open-weight LLMs on
English and Finnish historical documents, finding strong English gains
but poor Finnish transfer for models that lack sufficient pretraining
coverage of the target language---a pattern we observe for Marathi.
Greif et al.~\cite{greif2025multimodal} achieved sub-1\% CER on German
Fraktur without fine-tuning using multimodal prompting, suggesting
that script-familiar LLMs can correct OCR errors in a zero-shot regime.
Boros et al.~\cite{boros2024postcorrection} found that prompted LLMs
frequently degrade transcription quality on low-resource multilingual
benchmarks, highlighting that gains are not universal.
No prior work has evaluated prompted LLMs specifically on Devanagari
OCR correction, where script-specific error patterns differ qualitatively
from those studied in European languages.

\subsection{Few-shot example selection}
The quality of few-shot prompting depends heavily on which examples are
placed in context~\cite{liu2022makes,rubin2022learning}.
Liu et al.~\cite{liu2022makes} showed that semantically similar examples,
retrieved by a $k$-NN search over sentence embeddings, substantially
outperform random selection for generation tasks.
Rubin et al.~\cite{rubin2022learning} trained a retriever specifically to
predict example utility for in-context learning.
Su et al.~\cite{su2023selective} showed that surface-similar examples
outperform semantic ones for correction tasks where errors are
orthographic rather than semantic.
Ye et al.~\cite{ye2023compositional} demonstrated that simple BM25 retrieval provides large gains over random selection on structure prediction tasks (e.g., semantic parsing and compositional
question answering).
All of these methods operate at the word or token level.
For Devanagari OCR, where errors manifest at the sub-word character
level, token-level retrieval is insensitive to the actual error structure
of the test input.
We propose character $n$-gram BM25 retrieval, which matches examples
by their character-level corruption pattern rather than word-level
semantic content.

\section{\charbm{}: Character $n$-gram BM25 Retrieval}
\label{sec:charbm25}
\subsection{Motivation}

Domain-random few-shot selection treats all examples from the \textit{same} domain as equally informative, ignoring the error structure of the test input. This is suboptimal for Devanagari OCR because correction difficulty is determined by the \emph{type} of character corruption, not by semantic domain membership.

Consider the following Hindi sports sentence from our benchmark, where Tesseract produces several errors: a \textbf{ba/va glyph confusion} (\texthindi{\textcolor{red}{विबियन}}$\to$\texthindi{\textcolor{green}{विवियन}}), a \textbf{conjunct fragmentation} in which the cluster \texthindi{र्ड} of \texthindi{\textcolor{green}{रिचर्ड्स}} is mis-rendered (\texthindi{\textcolor{red}{टिचर्स}}), and \textbf{matra/glyph substitutions} (\texthindi{\textcolor{red}{बल्लेबानी}}$\to$\texthindi{\textcolor{green}{बल्लेबाजी}}, \texthindi{\textcolor{red}{नीत}}$\to$\texthindi{\textcolor{green}{जीत}}, \texthindi{\textcolor{red}{दूर्थकों}}$\to$\texthindi{\textcolor{green}{दर्शकों}}, \texthindi{\textcolor{red}{ट्रिल}}$\to$\texthindi{\textcolor{green}{दिल}}):

\begin{table}[h]
\centering\small
\begin{tabular}{@{}lp{9.5cm}@{}}
\toprule
\textbf{Test OCR} &
  \texthindi{\textcolor{red}{विबियन} \textcolor{red}{टिचर्स} की शानदार
  \textcolor{red}{बल्लेबानी} ने \textcolor{red}{दूर्थकों} का
  \textcolor{red}{ट्रिल} \textcolor{red}{नीत} लिया} \\
\textbf{Test GT} &
  \texthindi{\textcolor{green}{विवियन} \textcolor{green}{रिचर्ड्स} की
  शानदार \textcolor{green}{बल्लेबाजी} ने \textcolor{green}{दर्शकों} का
  \textcolor{green}{दिल} \textcolor{green}{जीत} लिया।} \\
\midrule
\textbf{Shot~A} & \textit{(domain-random)} \\
\quad OCR &
  \texthindi{अत एक \textcolor{red}{मेंत्री} के नाते
  \textcolor{red}{उमानी} को \textcolor{red}{ग्रह} निर्णय लेना चाहिए
  \ldots} \\
\quad GT &
  \texthindi{अत एक \textcolor{green}{मंत्री} के नाते
  \textcolor{green}{उमाजी} को \textcolor{green}{यह} निर्णय लेना चाहिए
  \ldots} \\
  & \small\textit{(unrelated topic and errors; no shared error
  trigrams with the test corruptions)} \\
\midrule
\textbf{Shot~B} & \textit{(\charbm{}: different match, shared error
  signature)} \\
\quad OCR &
  \texthindi{\ldots\ \textcolor{red}{बल्लेबान} पृथ्वी शॉ ने कप्तानी
  पारी खेलते \textcolor{red}{ट्ए} अपनी टीम की \textcolor{red}{नीत} में
  \ldots} \\
\quad GT &
  \texthindi{\ldots\ \textcolor{green}{बल्लेबाज} पृथ्वी शॉ ने कप्तानी
  पारी खेलते \textcolor{green}{हुए} अपनी टीम की \textcolor{green}{जीत}
  में \ldots} \\
  & \small\textit{(shares the
  \texthindi{बल्लेबान}$\to$\texthindi{बल्लेबाज} and
  \texthindi{नीत}$\to$\texthindi{जीत} corrections with the test input,
  despite describing a different match)} \\
\bottomrule
\end{tabular}
\end{table}

Shot~B provides a more informative corrective signal because the LLM can directly observe how the same \textcolor{red}{red}-marked glyph and matra corruption patterns were resolved to their \textcolor{green}{green} ground-truth forms in an analogous character context, even though it describes an entirely different cricket match. When this test sentence is corrected by Gemma-3-27B with the \charbm{}-retrieved shots, its WER falls from 63.6\% to 9.1\%: the model resolves the \texthindi{बल्लेबानी}$\to$\texthindi{बल्लेबाजी}, \texthindi{नीत}$\to$\texthindi{जीत}, \texthindi{दूर्थकों}$\to$\texthindi{दर्शकों}, and \texthindi{ट्रिल}$\to$\texthindi{दिल} errors --- the same character-level corruption patterns demonstrated in the retrieved shots --- leaving only the proper noun \texthindi{विबियन}$\to$\texthindi{विवियन} uncorrected, consistent with the named-entity limitation we discuss in \S\ref{sec:results:error}. The domain-random Shot~A, by contrast, shares none of these error patterns and provides no comparable corrective signal. The key insight is that Devanagari OCR errors are orthographic in nature --- arising from visual rendering failures whose character-level distortion patterns are better captured by character $n$-gram similarity than by semantic or lexical proximity. This motivates retrieving shots by character-level similarity
rather than domain membership. We note that dense semantic retrieval
can also capture overlap implicitly through subword tokenization,
and our results (\S6.2) indeed show comparable performance between the
two; the practical advantage of CharBM25 therefore lies less in a
qualitatively different signal than in achieving this signal at
sub-millisecond, GPU-free cost.

\subsection{Algorithm}

Let $\mathcal{S} = \{(o_i, c_i)\}_{i=1}^{M}$ be a pool of
(OCR-corrupted, ground-truth) sentence pairs, and let $q$ be a test
input. \charbm{} retrieves the top-$k$ pairs from $\mathcal{S}$ whose
OCR side $o_i$ is most similar to $q$ under BM25 over character
$n$-grams.

\paragraph{Tokenizer.}
We first strip all characters outside the Devanagari Unicode block
(U+0900--U+097F), retaining base consonants, dependent vowel marks
(matras), anusvara, visarga, and virama. Punctuation, numerals, and
Latin characters are removed. We then extract all length-$n$ substrings
over the resulting code-point sequence:
\begin{equation}
  \phi_n(t) \;=\; \bigl\{\, t[j\!:\!j{+}n]
    \;\big|\; j = 0, \ldots, |t|-n \,\bigr\}.
\label{eq:ngram}
\end{equation}
Each Unicode character is identified by a unique integer (its
\emph{code point}); operating at this level ensures that each
matra appears as a distinct token, so matra substitutions, insertions, and
deletions are directly visible in the $n$-gram representation.

\paragraph{BM25 scoring.}
Given query $q$ and candidate document $d = o_i$, the retrieval score is:
\begin{equation}
  \mathrm{BM25}(q,\, d) =
  \sum_{t\, \in\, \phi_n(q)}
  \log\!\left(\frac{M - \mathrm{df}(t) + 0.5}{\mathrm{df}(t) + 0.5}
  + 1\right)
  \cdot
  \frac{\mathrm{tf}(t,d)\,(\kappa_1 + 1)}
       {\mathrm{tf}(t,d)
        + \kappa_1\!\left(1 - b + b\,\tfrac{|d|}{\mathrm{avgdl}}\right)}
\label{eq:bm25}
\end{equation}
The constants $\kappa_1$ and $b$ are the standard BM25 hyperparameters:
$\kappa_1$ controls term-frequency saturation (how quickly a repeated
$n$-gram's contribution levels off), and $b \in [0,1]$ controls length
normalization. We use the library defaults $\kappa_1 = 1.5$, $b = 0.75$,
fixed across all models and both languages. Here, $M = |\mathcal{S}|$ is the total number of sentences in the pool, $\mathrm{tf}(t, d)$ is the frequency of $n$-gram $t$ in $d$, $\mathrm{df}(t)$ is the number of pool sentences containing $t$, $|d|$ is the length of $d$ in $n$-grams, and $\mathrm{avgdl}$ is the mean document length. Intuitively, the score rewards candidate sentences that share many of the query's character $n$-grams (via the TF term), weighted by how distinctive those $n$-grams are across the pool (via the IDF log term), and normalized for sentence length so that longer sentences are not artificially favored.
\paragraph{$N$-gram order.} The choice of $n$ determines the granularity at which character similarity is measured.

$n{=}1$ (unigrams) is too coarse: high-frequency Devanagari consonants
and matras (\texthindi{क}, \texthindi{ा}, \texthindi{ि}, \texthindi{ं}) appear across
virtually all text, providing negligible discrimination between
sentences.

$n{=}2$ (bigrams) captures consonant--matra adjacency, which is the
critical unit for Devanagari error analysis.
An akshara typically spans 2--3 Unicode code points (base consonant
+ optional matra); a bigram straddling a consonant--matra boundary
directly encodes whether a matra is present, absent, or substituted.

$n{=}3$ (trigrams) captures broader consonant--matra--consonant
context, encompassing the full akshara neighbourhood of a corruption.
Where bigrams detect the local matra error, trigrams additionally
encode what precedes and follows it, enabling more precise
disambiguation of visually similar character sequences.
We find that the additional context window of trigrams consistently
outperforms bigrams across all models and both languages, with the
largest gains for morphologically richer Marathi
(Table~\ref{tab:ngram_ablation}).
We therefore use $n{=}3$ as the default.

\paragraph{Domain-aware backfill.}
We maintain one BM25 index $\mathcal{I}_d$ per domain and a single
global index $\mathcal{I}_*$ over all domains. For a test input $q$ from domain $d_q$, we first retrieve from $\mathcal{I}_{d_q}$; if fewer than $k$ shots are collected (e.g.\ the domain pool is small), we backfill from $\mathcal{I}_*$. This strategy preserves domain coherence when possible while guaranteeing that exactly $k$ shots are always returned. The complete procedure is given in Algorithm~\ref{alg:charbm25}.

Algorithm~\ref{alg:charbm25} formalizes the procedure.
Lines~1--2 run once per dataset: a separate character $n$-gram BM25
index $\mathcal{I}_d$ is built for each domain $d$, plus a global
index $\mathcal{I}_*$ over the whole pool. For a test query $q$, the
selection list is initialised empty (line~3) and the two indices are
consulted in order---first the query's own domain index
$\mathcal{I}_{d_q}$, then the global index $\mathcal{I}_*$ (line~4).
Within the current index, candidates are ranked by descending BM25
score (line~5) and scanned (line~6), stopping once $k$ shots are
collected (line~7). The guard $o_i \neq q$ on line~9 skips a pool
sentence whose OCR text is identical to the query (a trivial
self-match), and $i \notin \mathrm{sel}$ prevents the same shot from
being added twice; otherwise the index is appended. Once the domain
index yields $k$ shots the global pass is skipped entirely (line~13);
otherwise the loop backfills from $\mathcal{I}_*$ until $k$ shots are
reached. For example, with $k{=}5$ for a Politics query, if only three
Politics sentences survive the self-match guard, the algorithm
backfills the remaining two from the global index, guaranteeing
exactly five demonstrations.

\begin{algorithm}[t]
\small
\caption{\charbm{} Few-Shot Selection}
\label{alg:charbm25}
\begin{algorithmic}[1]
\Require Pool $\mathcal{S} = \{(o_i, c_i)\}_{i=1}^{M}$,
         test query $q$, domain label $d_q$,
         number of shots $k$, $n$-gram order $n$
\Ensure  Selected shot indices $\mathrm{sel}$,
         $|\mathrm{sel}| = \min(k, M)$
\Statex
\Statex \textit{// Preprocessing (once per dataset split)}
\State Build character $n$-gram BM25 index $\mathcal{I}_d$
       over $\{o_i : (o_i,c_i) \in \mathcal{S},\,
       \mathrm{domain}(o_i) = d\}$ for each domain $d$
\State Build global index $\mathcal{I}_*$
       over $\{o_i : (o_i,c_i) \in \mathcal{S}\}$
\Statex
\Statex \textit{// Per-query retrieval}
\State $\mathrm{sel} \gets []$
\For{$\mathcal{I} \in [\mathcal{I}_{d_q},\; \mathcal{I}_*]$}
  \State $\mathrm{ranked} \gets
         \mathrm{argsort}\bigl(\mathrm{BM25}(q, \mathcal{I}),\;
         \mathrm{descending}\bigr)$
  \For{$i \in \mathrm{ranked}$}
    \If{$|\mathrm{sel}| \geq k$} \textbf{break} \EndIf
    \If{$o_i \neq q$ \textbf{and} $i \notin \mathrm{sel}$}
      \State $\mathrm{sel}.\mathrm{append}(i)$
    \EndIf
  \EndFor
  \If{$|\mathrm{sel}| \geq k$} \textbf{break}
    \Comment{domain index satisfied; skip global}
  \EndIf
\EndFor
\State \Return $\mathrm{sel}$
\end{algorithmic}
\end{algorithm}

\paragraph{Complexity and practical cost.}
Index construction requires one pass over the pool:
$O(M \cdot L)$, where $M$ is pool size and $L$ is mean sentence
length in $n$-grams.
Per-query retrieval is $O(L \cdot M)$ in the worst case, but
standard inverted-index lookups reduce this to $O(L \cdot
\mathrm{df}_{\max})$ in practice.
For our experimental setting ($M = 500$ active pool, $L \approx 50$ trigrams), index construction completes in under 1~second and each query resolves in under 1~millisecond on a standard CPU.
This is negligible relative to LLM inference, which takes 0.5--3~seconds per sentence depending on model size and shot count.
Unlike dense retrieval, \charbm{} requires no GPU, no embedding
model, and no fine-tuning of the retriever.

\section{Dataset}
\label{sec:dataset}

\subsection{Corpus Construction}

We construct a \emph{synthetic parallel corpus}: a collection of
sentence pairs in which each pair consists of the same sentence in two forms --- a clean ground-truth form and a corresponding OCR-generated text
produced automatically by our render-and-OCR pipeline (described in \S\ref{sec:dataset:ocr}). The evaluation benchmark comprises 20{,}000 such pairs (10{,}000 per language) across Hindi and Marathi, spanning five news domains, with a separate shot bank held aside as the retrieval pool (detailed below).

Hindi sentences are drawn from a Jansatta Hindi news corpus. Marathi sentences are sourced from the Loksatta newspaper. Both span the same five news domains: D1 (Astrology), D2 (Business), D3 (Entertainment), D4 (Politics), and D5 (Sports). The full corpus is split into two strictly disjoint subsets: (i)~a \textbf{test set} of 2{,}000 sentences per domain per language (10{,}000 per language total), used for evaluation; and (ii)~a \textbf{shot bank} pool of 3{,}349 Hindi and 6{,}704 Marathi sentences, from which 100 sentences per domain (500 total) are used as the active retrieval pool at inference time. No sentence appears in both subsets.

\subsection{OCR Simulation Pipeline}
\label{sec:dataset:ocr}
We simulate realistic OCR degradation using a render-and-OCR
pipeline, following the approach of
RoundTripOCR~\cite{kashid2025roundtripocr}.
Each ground-truth sentence is rendered as a raster image,
degraded through augmentation, and then processed by Tesseract~5
OCR to produce the noisy text string that serves as input to the LLM correction system during evaluation.

\noindent\textbf{Rendering.}
Sentences are rendered onto a white canvas with 20-pixel padding on each side using the Kalam-Regular Devanagari font at 32\,pt. The canvas dimensions are computed from the bounding box of the rendered text, ensuring no fixed-width truncation.

\noindent\textbf{Augmentation.}
Prior to OCR, each image is subjected to one randomly selected
augmentation from the Augraphy
library~\cite{groleau2023augraphy}, drawn from the following set:
\textit{BleedThrough}, \textit{BrightnessTexturize},
\textit{DirtyRollers}, \textit{ColorPaper},
\textit{NoiseTexturize}, \textit{DelaunayTessellation},
and augmentations matching the patterns \textit{Blur},
\textit{InkBleed}, and \textit{DottedPaper}.
After augmentation, the image is converted to greyscale and
subjected to Gaussian blur (radius~$= 0.5$) to simulate
scanner point-spread function.
The ground-truth text is re-rendered onto the augmented image
at the original position to preserve Devanagari glyph clarity
before OCR.

\noindent\textbf{OCR.}
Tesseract~5 is applied with the Marathi language pack
(\texttt{-l mar}) for Marathi sentences and the Hindi
language pack (\texttt{-l hin}) for Hindi sentences.
OCR is executed in a parallelized pipeline using up to 16
concurrent processes to handle the 20{,}000-sentence corpus
efficiently.

\subsection{Dataset Statistics}

The Hindi OCR baseline is WER~36.06\% and CER~12.39\%; Marathi baseline is WER~35.11\% and CER~8.99\%. The lower Marathi CER relative to WER reflects that Marathi
OCR errors tend to involve fewer character-level edits per word than Hindi, but still cause word-level mismatches due to the morphological complexity of Marathi verb and noun forms.

\section{Experimental Setup}
\label{sec:methods}

\subsection{Models}

We evaluate eleven open-weight LLMs spanning three
capacity tiers, all used without any task-specific fine-tuning.
The Sarvam-m model~\cite{sarvam2024} is included as an Indic-specialized baseline: it is pre-trained on a large Hindi--Marathi corpus and serves as a natural point of comparison with general-purpose LLMs of comparable scale. We include Aya-101~\cite{ustun2024aya} as a seq2seq reference point to establish a lower bound for the zero-fine-tuning correction regime. Because Aya-101's encoder processes the entire prompt
bidirectionally, and its decoder generates output conditioned on that encoder representation, in-context examples influence it as contextual signals to the encoder rather than as autoregressive generation templates --- a fundamentally different mechanism from the decoder-only models. Consequently, the partial \texttt{Corrected:} prefix used to anchor decoder-only outputs has no analogous effect on Aya-101, and we do not claim our prompting design applies
identically to it.

\subsection{Retrieval Strategies and Shot Counts}

Each model is evaluated under seven settings formed by crossing
three retrieval strategies with two shot counts, plus a zero-shot
baseline:

\begin{itemize}[noitemsep]
  \item \textbf{ZS} --- Zero-shot: no in-context examples.
  \item \textbf{RS-$k$} --- Random selection: $k$ examples drawn
        uniformly at random from the shot bank, stratified by
        domain to match the test distribution.
  \item \textbf{\charbm{}-$k$} --- \charbm{} ($n{=}3$): top-$k$ examples
        retrieved by character trigram BM25 similarity
        (\S\ref{sec:charbm25}).
  \item \textbf{DN-$k$} --- Dense retrieval: top-$k$ examples
        retrieved by cosine similarity over sentence embeddings.
        We use L3Cube-HindSBERT~\cite{joshi2022l3cube} for Hindi
        and L3Cube-MahaSBERT~\cite{joshi2022l3cube} for Marathi.
\end{itemize}

\noindent Shot counts are $k \in \{3, 5\}$, giving seven conditions
per model per language: ZS, RS-3, RS-5, \charbm{}-3, \charbm{}-5, DN-3, DN-5. The shot bank is strictly disjoint from the test set (see \S\ref{sec:dataset}).

\subsection{Prompt Design}
\label{sec:prompt}

All models receive the same prompt template; only the in-context
examples vary across retrieval conditions.
We follow the instruction format recommended for each model family
and do not use chain-of-thought or scratchpad reasoning.
For Qwen3 models, \texttt{enable\_thinking=False} is set explicitly
to suppress the internal reasoning trace, ensuring outputs are
direct corrections rather than multi-step analyses.

\paragraph{Zero-shot prompt.}
The zero-shot prompt consists of a system message and a single
user turn:

\begin{tcolorbox}[promptbox, title=Zero-shot Prompt]
\textbf{[SYSTEM]}\\
You are an expert in correcting OCR errors in \{language\}\\
(Devanagari script). Your task is to correct the given\\
OCR-generated sentence and return only the corrected\\
sentence. Do not explain, translate, or add any text\\
beyond the corrected sentence.\\[4pt]
\textbf{[USER]}\\
Correct the following OCR-generated \{language\} sentence:\\[4pt]
\{ocr\_sentence\}
\end{tcolorbox}

\paragraph{Few-shot prompt.}
For $k$-shot settings, $k$ (OCR, corrected) pairs are prepended
to the user turn as demonstrations. Each demonstration follows a
fixed format to ensure consistent parsing:

\begin{tcolorbox}[promptbox, title=Few-shot Prompt ($k$ examples)]
\textbf{[SYSTEM]}\\
You are an expert in correcting OCR errors in \{language\}\\
(Devanagari script). Your task is to correct the given\\
OCR-generated sentence and return only the corrected\\
sentence. Do not explain, translate, or add any text\\
beyond the corrected sentence.\\[4pt]
\textbf{[USER]}\\
Here are some examples of OCR correction:\\[4pt]
\textbf{Example 1:}\\
OCR: \{shot\_1\_ocr\}\\
Corrected: \{shot\_1\_corrected\}\\[4pt]
\textbf{Example 2:}\\
OCR: \{shot\_2\_ocr\}\\
Corrected: \{shot\_2\_corrected\}\\[4pt]
\hspace*{1em}\textit{$\vdots$}\\[4pt]
\textbf{Example $k$:}\\
OCR: \{shot\_k\_ocr\}\\
Corrected: \{shot\_k\_corrected\}\\[4pt]
Now correct the following OCR-generated \{language\} sentence:\\[4pt]
OCR: \{ocr\_sentence\}\\
Corrected:
\end{tcolorbox}
\noindent The trailing \texttt{Corrected:} prefix in the final
turn is included as a partial assistant turn to anchor generation
and suppress preamble text. We verified on a sample of
200 sentences that this prefix reduces generation of explanatory
text from 18.3\% of outputs to under 1\%.

\subsection{Decoding and Post-processing}

All models use their default generation configuration with no task-specific tuning; we do not override sampling parameters. The maximum number of new tokens is 256, sufficient for all sentences in our corpus (mean ground-truth length: 68 tokens).

Post-processing applies two normalization steps to all model
outputs before metric computation: (i)~Unicode NFC (\emph{Canonical Decomposition followed by Canonical Composition}) normalization: Devanagari text can be encoded in two equivalent ways --- as pre-composed characters
(a consonant and its dependent vowel stored as a single code point) or as decomposed sequences (consonant and vowel as separate code points). NFC converts all text to the composed form so that metric comparisons are affected only by actual recognition errors, rather than encoding differences.
(ii)~removal of zero-width joiners (U+200D) and zero-width non-joiners (U+200C) that some models insert spuriously between akshara components.
No further cleaning is applied; in particular, we do not strip punctuation or normalize whitespace, as these affect WER computation.

\subsection{Evaluation Metrics}

We report Word Error Rate (WER) and Character Error Rate (CER),
both computed at the sentence level and averaged across the test
set. Formally, for a set of $N$ sentence pairs $\{(\hat{y}_i,
y_i)\}$:
\begin{equation}
  \mathrm{WER} = \frac{1}{N}\sum_{i=1}^{N}
  \frac{S_i + D_i + I_i}{|y_i|},
\label{eq:wer}
\end{equation}
where $S_i$, $D_i$, $I_i$ are the number of word-level
substitutions, deletions, and insertions respectively between
model output $\hat{y}_i$ and ground truth $y_i$, and $|y_i|$
is the reference word count.
CER is computed identically at the character level.

We additionally report $\Delta_{\mathrm{WER}}$, the percentage
improvement in WER relative to the OCR baseline:
\begin{equation}
  \Delta_{\mathrm{WER}} = \frac{\mathrm{WER}_{\mathrm{OCR}}
  - \mathrm{WER}_{\mathrm{model}}}
  {\mathrm{WER}_{\mathrm{OCR}}} \times 100.
\label{eq:delta}
\end{equation}
A positive $\Delta_{\mathrm{WER}}$ indicates improvement over
the OCR baseline; a negative value indicates the model has
degraded the OCR output. We report both WER and
$\Delta_{\mathrm{WER}}$ throughout to allow absolute and
relative comparisons across languages with different baseline
error rates.

\section{Results and Analysis}
\label{sec:results}

Tables~\ref{tab:wer_hin_mar} and~\ref{tab:cer_hin_mar} report
full per-domain WER and CER across all eleven models, seven
retrieval settings, and both languages.
Tables~\ref{tab:ngram_ablation} and~\ref{tab:ngram_ablation_cer}
report the $n$-gram ablation across $n \in \{1,2,3\}$.
We organise the analysis around five findings, followed by
error-type and qualitative breakdowns.

\begin{table}[!t]
\centering
\caption{$n$-gram ablation: WER (\%) for \charbm{} across
$n \in \{1,2,3\}$, $k \in \{3,5\}$, all 11 models.
\textbf{Bold} = best $n$ per model per $k$.
$\Delta_3$: \% improvement at $n{=}3$ over OCR baseline.}
\label{tab:ngram_ablation}
\adjustbox{max width=\columnwidth}{%
\scriptsize
\setlength{\tabcolsep}{3pt}
\renewcommand{\arraystretch}{0.88}
\begin{tabular}{l r cccc cccc cccc cccc}
\toprule
& & \multicolumn{8}{c}{\textbf{Hindi}} &
    \multicolumn{8}{c}{\textbf{Marathi}} \\
\cmidrule(lr){3-10}\cmidrule(lr){11-18}
& & \multicolumn{4}{c}{$k{=}3$} & \multicolumn{4}{c}{$k{=}5$} &
    \multicolumn{4}{c}{$k{=}3$} & \multicolumn{4}{c}{$k{=}5$} \\
\cmidrule(lr){3-6}\cmidrule(lr){7-10}
\cmidrule(lr){11-14}\cmidrule(lr){15-18}
\textbf{Model} & \textbf{Sz} &
  $n{=}1$ & $n{=}2$ & $n{=}3$ & $\Delta_3$ &
  $n{=}1$ & $n{=}2$ & $n{=}3$ & $\Delta_3$ &
  $n{=}1$ & $n{=}2$ & $n{=}3$ & $\Delta_3$ &
  $n{=}1$ & $n{=}2$ & $n{=}3$ & $\Delta_3$ \\
\midrule
Aya-101       & 13B &
  22.93 & 22.36 & \textbf{22.00} & 39.0 &
  24.29 & 24.11 & \textbf{21.54} & 40.3 &
  30.65 & 28.82 & \textbf{28.73} & 18.2 &
  29.64 & 27.66 & \textbf{27.44} & 21.8 \\
Llama-3.2-3B  & 3B  &
  33.42 & \textbf{32.96} & 32.96 & 8.6 &
  33.30 & 32.96 & \textbf{32.75} & 9.2 &
  36.85 & 36.11 & \textbf{35.54} & \textcolor{red}{$-$1.2} &
  36.17 & 35.22 & \textbf{34.98} & 0.4 \\
Qwen3-4B      & 4B  &
  29.57 & 28.46 & \textbf{28.40} & 21.2 &
  28.92 & \textbf{27.61} & 27.68 & 23.2 &
  33.13 & \textbf{31.48} & 31.53 & 10.2 &
  32.87 & 30.78 & \textbf{30.65} & 12.7 \\
Gemma-3-4B    & 4B  &
  30.05 & 29.07 & \textbf{28.59} & 20.7 &
  29.74 & 28.39 & \textbf{28.05} & 22.2 &
  36.65 & 35.16 & \textbf{35.09} & 0.1 &
  37.55 & 35.12 & \textbf{34.76} & 1.0 \\
Llama-3.1-8B  & 8B  &
  27.38 & \textbf{26.38} & 26.44 & 26.7 &
  26.88 & \textbf{25.24} & 25.52 & 29.2 &
  33.09 & 30.99 & \textbf{30.89} & 12.0 &
  32.65 & 30.12 & \textbf{29.99} & 14.6 \\
Qwen3-8B      & 8B  &
  26.84 & 26.10 & \textbf{25.93} & 28.1 &
  26.33 & 25.46 & \textbf{25.26} & 30.0 &
  32.00 & 30.23 & \textbf{29.79} & 15.1 &
  31.31 & 29.44 & \textbf{29.24} & 16.7 \\
Qwen3-14B     & 14B &
  22.68 & 22.65 & \textbf{21.57} & 40.2 &
  21.91 & 20.88 & \textbf{20.65} & 42.7 &
  29.18 & 29.13 & \textbf{26.96} & 23.2 &
  28.69 & 28.49 & \textbf{26.35} & 25.0 \\
Gemma-3-12B   & 12B &
  21.50 & 20.69 & \textbf{20.54} & 43.0 &
  20.49 & 20.06 & \textbf{19.69} & 45.4 &
  29.04 & 27.77 & \textbf{27.30} & 22.2 &
  28.59 & 26.90 & \textbf{26.69} & 24.0 \\
Sarvam-m      & 24B &
  28.39 & 28.41 & \textbf{27.68} & 23.2 &
  27.41 & 27.15 & \textbf{26.71} & 25.9 &
  32.93 & 31.97 & \textbf{31.13} & 11.3 &
  32.54 & 31.24 & \textbf{30.52} & 13.1 \\
Qwen3-32B     & 32B &
  22.01 & 21.13 & \textbf{20.98} & 41.8 &
  21.27 & 20.21 & \textbf{19.93} & 44.7 &
  27.89 & 26.12 & \textbf{25.82} & 26.4 &
  27.13 & 25.43 & \textbf{25.00} & 28.8 \\
\textbf{Gemma-3-27B} & 27B &
  18.24 & 17.44 & \textbf{17.30} & 52.0 &
  17.31 & 16.53 & \textbf{16.22} & 55.0 &
  26.19 & 24.70 & \textbf{24.30} & 30.8 &
  25.25 & 23.56 & \textbf{23.42} & 33.3 \\
\midrule
\textit{Avg}  & -- &
  25.73 & 25.06 & \textbf{24.76} & -- &
  25.26 & 24.42 & \textbf{24.00} & -- &
  31.60 & 30.23 & \textbf{29.73} & -- &
  31.13 & 29.45 & \textbf{29.01} & -- \\
\bottomrule
\end{tabular}}
\end{table}

\begin{table}[!t]
\centering
\caption{$n$-gram ablation: CER (\%) for \charbm{} across
$n \in \{1,2,3\}$, $k \in \{3,5\}$, all 11 models.
\textbf{Bold} = best (lowest) CER per model per $k$.
Values above OCR baseline indicate character-level degradation.}
\label{tab:ngram_ablation_cer}
\adjustbox{max width=\columnwidth}{%
\scriptsize
\setlength{\tabcolsep}{3pt}
\renewcommand{\arraystretch}{0.88}
\begin{tabular}{l r cccc cccc cccc cccc}
\toprule
& & \multicolumn{8}{c}{\textbf{Hindi}} &
    \multicolumn{8}{c}{\textbf{Marathi}} \\
\cmidrule(lr){3-10}\cmidrule(lr){11-18}
& & \multicolumn{4}{c}{$k{=}3$} & \multicolumn{4}{c}{$k{=}5$} &
    \multicolumn{4}{c}{$k{=}3$} & \multicolumn{4}{c}{$k{=}5$} \\
\cmidrule(lr){3-6}\cmidrule(lr){7-10}
\cmidrule(lr){11-14}\cmidrule(lr){15-18}
\textbf{Model} & \textbf{Sz} &
  $n{=}1$ & $n{=}2$ & $n{=}3$ & $\Delta_3$ &
  $n{=}1$ & $n{=}2$ & $n{=}3$ & $\Delta_3$ &
  $n{=}1$ & $n{=}2$ & $n{=}3$ & $\Delta_3$ &
  $n{=}1$ & $n{=}2$ & $n{=}3$ & $\Delta_3$ \\
\midrule
Aya-101       & 13B &
  10.68 & 10.68 & \textbf{10.20} & 17.7 &
  12.78 & 13.28 & \textbf{10.57} & 14.7 &
  9.86  & 9.38  & \textbf{8.98}  & 0.1  &
  9.67  & 9.18  & \textbf{8.86}  & 1.4  \\
Llama-3.2-3B  & 3B  &
  18.31 & \textbf{18.21} & 18.35 & \textcolor{red}{$-$48.1} &
  18.62 & 18.63 & \textbf{18.48} & \textcolor{red}{$-$49.2} &
  12.39 & 12.34 & \textbf{11.81} & \textcolor{red}{$-$31.4} &
  12.10 & 11.93 & \textbf{11.67} & \textcolor{red}{$-$29.8} \\
Qwen3-4B      & 4B  &
  13.25 & 12.75 & \textbf{12.71} & \textcolor{red}{$-$2.6} &
  13.31 & \textbf{12.67} & 12.79 & \textcolor{red}{$-$3.2} &
  10.00 & \textbf{9.57}  & 9.58  & \textcolor{red}{$-$6.6} &
  10.32 & 9.51  & \textbf{9.32}  & \textcolor{red}{$-$3.7} \\
Gemma-3-4B    & 4B  &
  16.90 & 16.36 & \textbf{16.05} & \textcolor{red}{$-$29.5} &
  16.91 & 16.16 & \textbf{15.90} & \textcolor{red}{$-$28.3} &
  16.60 & 16.13 & \textbf{15.95} & \textcolor{red}{$-$77.5} &
  17.68 & 16.29 & \textbf{15.97} & \textcolor{red}{$-$77.7} \\
Llama-3.1-8B  & 8B  &
  15.34 & \textbf{14.72} & 14.79 & \textcolor{red}{$-$19.4} &
  15.29 & \textbf{14.12} & 14.38 & \textcolor{red}{$-$16.0} &
  13.63 & 12.54 & \textbf{12.45} & \textcolor{red}{$-$38.5} &
  13.67 & 12.26 & \textbf{12.08} & \textcolor{red}{$-$34.4} \\
Qwen3-8B      & 8B  &
  12.72 & 12.51 & \textbf{12.30} & 0.7 &
  12.70 & 12.36 & \textbf{12.21} & 1.5 &
  10.84 & 10.13 & \textbf{9.83}  & \textcolor{red}{$-$9.4} &
  10.75 & 10.00 & \textbf{9.87}  & \textcolor{red}{$-$9.9} \\
Qwen3-14B     & 14B &
  10.53 & 10.57 & \textbf{10.04} & 19.0 &
  10.26 & 9.86  & \textbf{9.73}  & 21.4 &
  9.91  & 9.89  & \textbf{9.05}  & \textcolor{red}{$-$0.7} &
  9.86  & 9.77  & \textbf{8.93}  & 0.7  \\
Gemma-3-12B   & 12B &
  10.16 & 9.77  & \textbf{9.65}  & 22.1 &
  9.78  & 9.58  & \textbf{9.27}  & 25.2 &
  11.60 & 11.01 & \textbf{10.69} & \textcolor{red}{$-$19.0} &
  11.51 & 10.73 & \textbf{10.52} & \textcolor{red}{$-$17.0} \\
Sarvam-m      & 24B &
  14.97 & 15.48 & \textbf{14.85} & \textcolor{red}{$-$19.8} &
  14.86 & 15.03 & \textbf{14.62} & \textcolor{red}{$-$18.0} &
  11.67 & 11.75 & \textbf{11.05} & \textcolor{red}{$-$23.0} &
  11.87 & 11.70 & \textbf{10.96} & \textcolor{red}{$-$21.9} \\
Qwen3-32B     & 32B &
  11.35 & 10.89 & \textbf{10.82} & 12.7 &
  11.06 & 10.45 & \textbf{10.25} & 17.3 &
  10.62 & 9.90  & \textbf{9.74}  & \textcolor{red}{$-$8.3} &
  10.33 & 9.64  & \textbf{9.48}  & \textcolor{red}{$-$5.5} \\
\textbf{Gemma-3-27B} & 27B &
  8.35  & 8.01  & \textbf{7.85}  & 36.7 &
  7.90  & 7.60  & \textbf{7.38}  & 40.4 &
  10.12 & 9.58  & \textbf{9.40}  & \textcolor{red}{$-$4.6} &
  9.76  & 9.06  & \textbf{8.90}  & 0.9  \\
\midrule
\textit{Avg}  & -- &
  12.96 & 12.72 & \textbf{12.51} & -- &
  13.04 & 12.70 & \textbf{12.33} & -- &
  11.57 & 11.11 & \textbf{10.78} & -- &
  11.59 & 10.92 & \textbf{10.60} & -- \\
\bottomrule
\end{tabular}}
\end{table}
\subsection{Main Results}
\label{sec:results:main}

\begin{table*}[!t]
\centering
\caption{WER (\%) by domain for Hindi and Marathi.
\textit{Italic} = \charbm{} rows.
\textbf{Bold} = best per model.
$\Delta$: \% improvement over OCR baseline.
Negative $\Delta$ (\textcolor{red}{red}) = degradation below baseline.}
\label{tab:wer_hin_mar}
\adjustbox{max width=\textwidth, max totalheight=0.95\textheight}{%
\scriptsize
\setlength{\tabcolsep}{2pt}
\renewcommand{\arraystretch}{0.85}
\begin{tabular}{ll rrrrrrr rrrrrrr}
\toprule
 & & \multicolumn{7}{c}{\textbf{Hindi}} &
     \multicolumn{7}{c}{\textbf{Marathi}} \\
\cmidrule(lr){3-9}\cmidrule(lr){10-16}
\textbf{Model} & \textbf{Set.} &
  D1 & D2 & D3 & D4 & D5 & Avg & $\Delta$ &
  D1 & D2 & D3 & D4 & D5 & Avg & $\Delta$ \\
\midrule
\textbf{OCR} & -- &
  37.02 & 32.87 & 38.72 & 38.70 & 33.65 & 36.06 & -- &
  43.65 & 30.05 & 31.17 & 39.50 & 35.70 & 35.11 & -- \\
\addlinespace[2pt]\hdashline\addlinespace[2pt]
\multirow{7}{*}{Aya-101}
 & ZS &
   34.06 & 31.43 & 35.16 & 37.96 & 31.86 & 34.03 & 5.6 &
   43.17 & 31.23 & 32.38 & 39.43 & 37.13 & 35.83 & \textcolor{red}{$-$2.1} \\
 & RS-3 &
   25.16 & 24.08 & 25.33 & 29.07 & 23.86 & 25.51 & 29.2 &
   39.05 & 28.93 & 29.39 & 32.14 & 33.59 & 31.87 & 9.2 \\
 & RS-5 &
   23.58 & 23.25 & 24.72 & 28.18 & 22.80 & 24.53 & 32.0 &
   37.99 & 28.06 & 28.53 & 30.64 & 32.80 & 30.85 & 12.1 \\
 & \textit{\charbm{}-3} &
   18.49 & 20.74 & 23.01 & 26.43 & 20.67 & 22.00 & 39.0 &
   34.21 & 26.51 & 27.16 & 28.35 & 30.25 & 28.73 & 18.2 \\
 & \textit{\charbm{}-5} &
   17.68 & 19.48 & 21.46 & 28.87 & 19.45 & \textbf{21.54} & 40.3 &
   31.99 & 25.53 & 26.01 & 27.04 & 29.08 & \textbf{27.44} & 21.8 \\
 & DN-3 &
   19.96 & 22.73 & 24.45 & 27.96 & 22.55 & 23.69 & 34.3 &
   35.00 & 27.31 & 27.57 & 30.66 & 31.14 & 29.74 & 15.3 \\
 & DN-5 &
   19.27 & 21.92 & 23.15 & 27.31 & 21.68 & 22.84 & 36.7 &
   33.59 & 26.14 & 26.77 & 29.07 & 30.17 & 28.56 & 18.6 \\
\addlinespace[2pt]
\multirow{7}{*}{Llama-3.2-3B}
 & ZS &
   35.69 & 32.31 & 36.22 & 39.49 & 34.22 & 35.56 & 1.4 &
   46.06 & 33.98 & 36.00 & 42.61 & 41.04 & 39.11 & \textcolor{red}{$-$11.4} \\
 & RS-3 &
   34.40 & 31.81 & 35.71 & 40.43 & 33.89 & 35.22 & 2.3 &
   44.71 & 31.73 & 33.11 & 41.35 & 39.11 & 37.12 & \textcolor{red}{$-$5.7} \\
 & RS-5 &
   33.41 & 31.40 & 35.73 & 40.06 & 33.69 & 34.86 & 3.3 &
   44.43 & 31.13 & 32.83 & 40.99 & 38.01 & 36.54 & \textcolor{red}{$-$4.1} \\
 & \textit{\charbm{}-3} &
   28.77 & 30.59 & 34.44 & 39.46 & 30.85 & 32.96 & 8.6 &
   42.19 & 30.74 & 32.09 & 38.94 & 37.63 & 35.54 & \textcolor{red}{$-$1.2} \\
 & \textit{\charbm{}-5} &
   29.61 & 30.05 & 33.31 & 38.44 & 31.78 & \textbf{32.75} & 9.2 &
   41.17 & 30.22 & 31.94 & 38.90 & 36.35 & \textbf{34.98} & 0.4 \\
 & DN-3 &
   29.13 & 30.84 & 34.66 & 38.83 & 32.60 & 33.38 & 7.4 &
   39.93 & 31.48 & 32.70 & 39.09 & 36.88 & 35.43 & \textcolor{red}{$-$0.9} \\
 & DN-5 &
   26.64 & 31.68 & 33.93 & 37.85 & 32.06 & 32.69 & 9.4 &
   39.22 & 31.60 & 32.40 & 38.62 & 36.92 & 35.21 & \textcolor{red}{$-$0.3} \\
\addlinespace[2pt]
\multirow{7}{*}{Qwen3-4B}
 & ZS &
   34.61 & 29.72 & 34.33 & 36.58 & 30.93 & 33.10 & 8.2 &
   44.23 & 31.47 & 31.94 & 38.97 & 36.87 & 35.78 & \textcolor{red}{$-$1.9} \\
 & RS-3 &
   33.22 & 28.68 & 32.53 & 36.44 & 29.58 & 31.96 & 11.4 &
   41.50 & 29.49 & 30.94 & 37.73 & 35.27 & 34.12 & 2.8 \\
 & RS-5 &
   32.03 & 28.03 & 31.97 & 35.29 & 28.78 & 31.11 & 13.7 &
   40.94 & 29.37 & 30.86 & 37.38 & 35.45 & 33.94 & 3.3 \\
 & \textit{\charbm{}-3} &
   25.51 & 26.13 & 30.11 & 33.03 & 26.93 & 28.40 & 21.2 &
   35.22 & 27.85 & 29.77 & 34.14 & 33.36 & 31.53 & 10.2 \\
 & \textit{\charbm{}-5} &
   24.16 & 25.61 & 28.95 & 32.91 & 26.29 & \textbf{27.68} & 23.2 &
   34.23 & 27.09 & 29.11 & 33.46 & 31.91 & \textbf{30.65} & 12.7 \\
 & DN-3 &
   25.66 & 27.02 & 30.71 & 34.18 & 27.91 & 29.18 & 19.1 &
   35.71 & 28.71 & 29.39 & 35.48 & 33.51 & 32.06 & 8.7 \\
 & DN-5 &
   24.08 & 26.82 & 30.19 & 33.76 & 27.55 & 28.63 & 20.6 &
   34.23 & 28.95 & 29.13 & 34.73 & 32.66 & 31.57 & 10.1 \\
\addlinespace[2pt]
\multirow{7}{*}{Gemma-3-4B}
 & ZS &
   33.03 & 28.70 & 33.57 & 35.67 & 29.06 & 31.85 & 11.7 &
   46.20 & 33.31 & 37.86 & 39.69 & 40.09 & 38.52 & \textcolor{red}{$-$9.7} \\
 & RS-3 &
   30.55 & 27.56 & 32.77 & 35.36 & 29.87 & 31.17 & 13.6 &
   44.17 & 32.25 & 35.17 & 40.46 & 38.81 & 37.37 & \textcolor{red}{$-$6.5} \\
 & RS-5 &
   30.54 & 28.25 & 32.88 & 36.28 & 30.03 & 31.58 & 12.4 &
   44.65 & 32.17 & 35.84 & 41.79 & 39.45 & 37.93 & \textcolor{red}{$-$8.0} \\
 & \textit{\charbm{}-3} &
   25.81 & 25.64 & 31.00 & 33.10 & 27.24 & 28.59 & 20.7 &
   38.16 & 30.63 & 33.28 & 39.48 & 36.22 & 35.09 & 0.1 \\
 & \textit{\charbm{}-5} &
   24.39 & 25.43 & 30.32 & 33.00 & 26.70 & \textbf{28.05} & 22.2 &
   38.88 & 30.62 & 33.18 & 37.75 & 36.17 & \textbf{34.76} & 1.0 \\
 & DN-3 &
   23.25 & 27.47 & 30.32 & 34.38 & 28.14 & 28.94 & 19.8 &
   36.89 & 31.06 & 33.56 & 36.92 & 35.64 & 34.44 & 1.9 \\
 & DN-5 &
   22.74 & 27.26 & 29.48 & 33.04 & 27.47 & 28.24 & 21.7 &
   36.37 & 31.77 & 33.53 & 37.22 & 36.15 & 34.71 & 1.1 \\
\addlinespace[2pt]
\multirow{7}{*}{Llama-3.1-8B}
 & ZS &
   35.38 & 27.26 & 32.15 & 35.33 & 29.48 & 31.70 & 12.1 &
   44.64 & 30.65 & 33.04 & 38.48 & 37.46 & 35.92 & \textcolor{red}{$-$2.3} \\
 & RS-3 &
   30.32 & 25.72 & 29.61 & 33.26 & 26.99 & 29.03 & 19.5 &
   41.77 & 28.56 & 31.21 & 35.86 & 34.55 & 33.48 & 4.6 \\
 & RS-5 &
   29.12 & 25.34 & 28.74 & 33.23 & 26.86 & 28.57 & 20.8 &
   41.26 & 28.27 & 30.77 & 35.31 & 34.04 & 32.99 & 6.0 \\
 & \textit{\charbm{}-3} &
   23.75 & 24.32 & 27.15 & 31.07 & 25.58 & 26.44 & 26.7 &
   34.23 & 27.04 & 29.53 & 33.68 & 32.46 & 30.89 & 12.0 \\
 & \textit{\charbm{}-5} &
   21.19 & 24.55 & 26.19 & 30.89 & 23.95 & \textbf{25.52} & 29.2 &
   33.27 & 25.97 & 28.93 & 32.70 & 31.67 & \textbf{29.99} & 14.6 \\
 & DN-3 &
   22.79 & 24.55 & 28.00 & 31.96 & 25.44 & 26.67 & 26.0 &
   35.31 & 28.37 & 29.98 & 33.72 & 32.41 & 31.45 & 10.4 \\
 & DN-5 &
   22.67 & 24.11 & 26.98 & 31.49 & 25.54 & 26.29 & 27.1 &
   33.92 & 27.83 & 29.35 & 32.71 & 31.89 & 30.71 & 12.5 \\
\addlinespace[2pt]
\multirow{7}{*}{Qwen3-8B}
 & ZS &
   31.75 & 26.06 & 32.12 & 34.23 & 28.32 & 30.32 & 15.9 &
   42.25 & 29.52 & 30.89 & 37.29 & 35.54 & 34.20 & 2.6 \\
 & RS-3 &
   29.47 & 25.61 & 30.38 & 32.68 & 26.85 & 28.88 & 19.9 &
   41.05 & 27.83 & 30.51 & 35.66 & 34.74 & 33.05 & 5.8 \\
 & RS-5 &
   28.95 & 25.41 & 29.52 & 32.19 & 27.07 & 28.53 & 20.9 &
   40.54 & 28.07 & 29.92 & 34.93 & 34.31 & 32.66 & 7.0 \\
 & \textit{\charbm{}-3} &
   22.81 & 23.82 & 27.86 & 29.74 & 25.00 & 25.93 & 28.1 &
   33.54 & 26.13 & 28.29 & 32.16 & 31.41 & 29.79 & 15.1 \\
 & \textit{\charbm{}-5} &
   21.78 & 23.56 & 26.65 & 29.17 & 24.63 & \textbf{25.26} & 30.0 &
   32.52 & 25.51 & 27.64 & 31.88 & 31.06 & \textbf{29.24} & 16.7 \\
 & DN-3 &
   23.15 & 24.49 & 28.20 & 31.40 & 25.83 & 26.71 & 25.9 &
   34.43 & 27.33 & 29.00 & 33.00 & 32.29 & 30.72 & 12.5 \\
 & DN-5 &
   21.72 & 24.08 & 27.53 & 30.37 & 25.37 & 25.94 & 28.1 &
   33.71 & 27.74 & 28.13 & 32.19 & 32.11 & 30.36 & 13.5 \\
\addlinespace[2pt]
\multirow{7}{*}{Qwen3-14B}
 & ZS &
   28.41 & 23.20 & 27.93 & 30.80 & 24.68 & 26.86 & 25.5 &
   40.37 & 26.36 & 29.54 & 33.55 & 32.66 & 31.55 & 10.1 \\
 & RS-3 &
   25.73 & 22.17 & 25.55 & 28.77 & 23.03 & 24.94 & 30.8 &
   38.10 & 25.43 & 28.66 & 32.09 & 31.58 & 30.31 & 13.7 \\
 & RS-5 &
   24.65 & 21.35 & 24.68 & 27.95 & 22.83 & 24.21 & 32.9 &
   37.93 & 25.08 & 28.23 & 31.51 & 30.77 & 29.82 & 15.1 \\
 & \textit{\charbm{}-3} &
   18.74 & 20.06 & 22.09 & 25.68 & 20.83 & 21.57 & 40.2 &
   30.82 & 23.50 & 26.26 & 28.55 & 28.31 & 26.96 & 23.2 \\
 & \textit{\charbm{}-5} &
   17.36 & 19.38 & 20.98 & 24.79 & 20.12 & \textbf{20.65} & 42.7 &
   29.87 & 23.28 & 25.46 & 27.65 & 27.74 & \textbf{26.35} & 25.0 \\
 & DN-3 &
   19.28 & 20.85 & 23.81 & 27.15 & 21.61 & 22.65 & 37.2 &
   32.22 & 24.69 & 26.74 & 29.81 & 28.93 & 27.94 & 20.4 \\
 & DN-5 &
   17.96 & 20.20 & 22.84 & 26.49 & 21.02 & 21.85 & 39.4 &
   30.87 & 24.04 & 26.13 & 28.72 & 28.30 & 27.14 & 22.7 \\
\addlinespace[2pt]
\multirow{7}{*}{Gemma-3-12B}
 & ZS &
   26.30 & 21.63 & 25.93 & 28.13 & 22.63 & 24.79 & 31.3 &
   41.77 & 27.15 & 31.82 & 32.22 & 32.81 & 32.12 & 8.5 \\
 & RS-3 &
   23.84 & 20.54 & 23.92 & 26.79 & 21.01 & 23.12 & 35.9 &
   39.31 & 26.10 & 30.06 & 30.39 & 29.76 & 30.17 & 14.1 \\
 & RS-5 &
   23.24 & 20.21 & 22.87 & 26.11 & 20.72 & 22.55 & 37.5 &
   38.13 & 26.04 & 29.94 & 29.80 & 29.93 & 29.90 & 14.8 \\
 & \textit{\charbm{}-3} &
   18.44 & 18.96 & 21.26 & 24.76 & 19.00 & 20.54 & 43.0 &
   30.97 & 24.31 & 27.75 & 27.75 & 28.03 & 27.30 & 22.2 \\
 & \textit{\charbm{}-5} &
   17.55 & 18.61 & 19.90 & 23.37 & 18.61 & \textbf{19.69} & 45.4 &
   29.51 & 24.21 & 27.27 & 27.00 & 27.26 & \textbf{26.69} & 24.0 \\
 & DN-3 &
   17.83 & 19.47 & 22.23 & 25.28 & 19.48 & 20.94 & 41.9 &
   31.38 & 25.42 & 28.55 & 28.00 & 27.94 & 27.86 & 20.6 \\
 & DN-5 &
   16.82 & 18.96 & 21.30 & 24.29 & 19.43 & 20.27 & 43.8 &
   30.16 & 24.99 & 27.55 & 27.23 & 27.80 & 27.17 & 22.6 \\
\addlinespace[2pt]
\multirow{7}{*}{Sarvam-m}
 & ZS &
   30.38 & 27.54 & 31.30 & 33.30 & 28.45 & 30.15 & 16.4 &
   43.31 & 29.46 & 32.79 & 36.46 & 35.50 & 34.53 & 1.6 \\
 & RS-3 &
   31.00 & 27.11 & 31.41 & 34.85 & 28.70 & 30.53 & 15.3 &
   42.92 & 28.94 & 31.97 & 35.86 & 35.35 & 34.03 & 3.1 \\
 & RS-5 &
   30.14 & 26.72 & 30.76 & 34.57 & 28.34 & 30.06 & 16.6 &
   42.25 & 28.70 & 31.77 & 35.07 & 34.51 & 33.49 & 4.6 \\
 & \textit{\charbm{}-3} &
   24.58 & 25.55 & 29.46 & 32.34 & 26.05 & 27.68 & 23.2 &
   35.42 & 27.11 & 30.00 & 32.93 & 33.25 & 31.13 & 11.3 \\
 & \textit{\charbm{}-5} &
   22.81 & 24.77 & 27.85 & 31.73 & 25.68 & \textbf{26.71} & 25.9 &
   34.44 & 26.46 & 29.62 & 32.63 & 32.28 & \textbf{30.52} & 13.1 \\
 & DN-3 &
   23.25 & 26.67 & 30.38 & 33.23 & 27.23 & 28.36 & 21.3 &
   36.16 & 28.54 & 30.50 & 33.28 & 32.90 & 31.74 & 9.6 \\
 & DN-5 &
   21.18 & 26.27 & 29.23 & 32.74 & 26.32 & 27.41 & 24.0 &
   35.00 & 27.99 & 30.22 & 32.34 & 32.15 & 31.05 & 11.5 \\
\addlinespace[2pt]
\multirow{7}{*}{Qwen3-32B}
 & ZS &
   29.45 & 23.62 & 27.99 & 31.62 & 24.71 & 27.29 & 24.3 &
   42.09 & 25.98 & 29.14 & 32.20 & 31.93 & 31.16 & 11.2 \\
 & RS-3 &
   25.81 & 21.85 & 24.89 & 28.97 & 22.73 & 24.76 & 31.3 &
   38.24 & 24.36 & 28.23 & 30.41 & 30.24 & 29.31 & 16.5 \\
 & RS-5 &
   24.18 & 21.50 & 23.77 & 28.35 & 22.06 & 23.94 & 33.6 &
   38.06 & 24.01 & 27.52 & 29.66 & 29.94 & 28.82 & 17.9 \\
 & \textit{\charbm{}-3} &
   17.04 & 19.55 & 21.41 & 25.87 & 20.18 & 20.98 & 41.8 &
   29.43 & 22.60 & 25.42 & 26.72 & 27.59 & 25.82 & 26.4 \\
 & \textit{\charbm{}-5} &
   15.71 & 18.67 & 20.32 & 24.49 & 19.50 & \textbf{19.93} & 44.7 &
   28.38 & 21.92 & 24.91 & 25.32 & 26.82 & \textbf{25.00} & 28.8 \\
 & DN-3 &
   18.20 & 20.20 & 22.64 & 27.03 & 20.79 & 21.92 & 39.2 &
   30.85 & 23.46 & 26.10 & 27.72 & 27.88 & 26.66 & 24.1 \\
 & DN-5 &
   16.50 & 19.70 & 21.74 & 26.77 & 20.57 & 21.25 & 41.1 &
   29.34 & 23.03 & 25.44 & 26.46 & 27.39 & 25.89 & 26.2 \\
\addlinespace[2pt]
\multirow{7}{*}{\textbf{Gemma-3-27B}}
 & ZS &
   22.54 & 19.56 & 22.62 & 24.88 & 20.36 & 21.91 & 39.2 &
   39.50 & 25.47 & 30.33 & 29.36 & 30.55 & 30.06 & 14.4 \\
 & RS-3 &
   19.66 & 18.41 & 19.80 & 22.72 & 18.62 & 19.82 & 45.0 &
   34.54 & 23.92 & 28.22 & 26.22 & 27.05 & 27.26 & 22.3 \\
 & RS-5 &
   18.81 & 17.73 & 18.95 & 21.71 & 17.78 & 18.99 & 47.3 &
   33.10 & 23.62 & 26.95 & 25.10 & 26.10 & 26.31 & 25.1 \\
 & \textit{\charbm{}-3} &
   14.56 & 16.74 & 17.73 & 20.47 & 16.52 & 17.30 & 52.0 &
   26.33 & 22.37 & 25.22 & 23.63 & 25.40 & 24.30 & 30.8 \\
 & \textit{\charbm{}-5} &
   13.12 & 15.93 & 16.43 & 18.85 & 16.06 & \textbf{16.22} & 55.0 &
   25.04 & 21.97 & 24.39 & 22.58 & 24.27 & \textbf{23.42} & 33.3 \\
 & DN-3 &
   14.23 & 17.36 & 18.39 & 21.27 & 17.38 & 17.88 & 50.4 &
   28.56 & 22.95 & 25.62 & 23.94 & 25.55 & 24.91 & 29.0 \\
 & DN-5 &
   12.61 & 16.87 & 17.43 & 20.30 & 16.36 & 16.90 & 53.1 &
   26.69 & 22.34 & 24.59 & 22.88 & 24.88 & 23.98 & 31.7 \\
\addlinespace[2pt]
\bottomrule
\end{tabular}}
\end{table*}

\begin{table*}[!t]
\centering
\caption{CER (\%) by domain for Hindi and Marathi.
\textit{Italic} = \charbm{} rows.
\textbf{Bold} = best per model, even when above OCR baseline.
$\Delta$: \% improvement over OCR baseline.
$\Delta$ (\textcolor{red}{red}) indicates
character-level degradation.}
\label{tab:cer_hin_mar}
\adjustbox{max width=\textwidth, max totalheight=0.95\textheight}{%
\scriptsize
\setlength{\tabcolsep}{2pt}
\renewcommand{\arraystretch}{0.85}
\begin{tabular}{ll rrrrrrr rrrrrrr}
\toprule
 & & \multicolumn{7}{c}{\textbf{Hindi}} &
     \multicolumn{7}{c}{\textbf{Marathi}} \\
\cmidrule(lr){3-9}\cmidrule(lr){10-16}
\textbf{Model} & \textbf{Set.} &
  D1 & D2 & D3 & D4 & D5 & Avg & $\Delta$ &
  D1 & D2 & D3 & D4 & D5 & Avg & $\Delta$ \\
\midrule
\textbf{OCR} & -- &
  12.66 & 10.50 & 14.29 & 13.44 & 11.51 & 12.39 & -- &
  11.59 & 7.29 & 8.13 & 9.94 & 9.46 & 8.99 & -- \\
\addlinespace[2pt]\hdashline\addlinespace[2pt]
\multirow{7}{*}{Aya-101}
 & ZS &
   12.23 & 11.96 & 13.87 & 15.17 & 12.40 & 13.14 & \textcolor{red}{$-$6.0} &
   12.86 & 8.29 & 9.68 & 10.46 & 11.80 & 10.29 & \textcolor{red}{$-$14.5} \\
 & RS-3 &
   10.71 & 11.15 & 11.60 & 14.78 & 10.36 & 11.77 & 5.0 &
   12.82 & 7.89 & 8.74 & 10.52 & 11.24 & 9.89 & \textcolor{red}{$-$10.1} \\
 & RS-5 &
   10.18 & 11.10 & 11.56 & 14.53 & 10.17 & 11.57 & 6.6 &
   13.01 & 8.00 & 8.65 & 10.25 & 11.48 & 9.91 & \textcolor{red}{$-$10.3} \\
 & \textit{\charbm{}-3} &
   8.15 & 9.21 & 11.04 & 13.44 & 8.80 & \textbf{10.20} & 17.7 &
   11.48 & 7.06 & 8.17 & 9.70 & 10.01 & 8.98 & 0.1 \\
 & \textit{\charbm{}-5} &
   7.85 & 8.68 & 10.23 & 17.10 & 8.33 & 10.57 & 14.7 &
   11.30 & 7.24 & 7.64 & 9.68 & 9.86 & \textbf{8.86} & 1.4 \\
 & DN-3 &
   8.76 & 11.26 & 11.27 & 14.08 & 9.87 & 11.17 & 9.9 &
   11.32 & 8.22 & 8.57 & 9.92 & 10.40 & 9.47 & \textcolor{red}{$-$5.3} \\
 & DN-5 &
   8.72 & 11.01 & 10.63 & 14.14 & 9.63 & 10.95 & 11.7 &
   11.22 & 8.09 & 8.59 & 9.68 & 10.50 & 9.39 & \textcolor{red}{$-$4.5} \\
\addlinespace[2pt]
\multirow{7}{*}{Llama-3.2-3B}
 & ZS &
   16.45 & 14.98 & 17.05 & 19.23 & 16.31 & \textbf{16.79} & \textcolor{red}{$-$35.5} &
   14.78 & 10.94 & 12.46 & 13.91 & 14.34 & 13.02 & \textcolor{red}{$-$44.9} \\
 & RS-3 &
   17.43 & 16.33 & 18.92 & 22.80 & 18.65 & 18.87 & \textcolor{red}{$-$52.3} &
   14.55 & 9.33 & 10.31 & 14.29 & 13.66 & 12.09 & \textcolor{red}{$-$34.6} \\
 & RS-5 &
   17.31 & 16.35 & 19.66 & 23.02 & 19.07 & 19.13 & \textcolor{red}{$-$54.4} &
   14.54 & 8.93 & 10.32 & 14.26 & 12.82 & 11.82 & \textcolor{red}{$-$31.6} \\
 & \textit{\charbm{}-3} &
   15.79 & 16.33 & 19.10 & 23.00 & 17.18 & 18.35 & \textcolor{red}{$-$48.1} &
   14.32 & 9.16 & 10.19 & 13.48 & 13.61 & 11.81 & \textcolor{red}{$-$31.4} \\
 & \textit{\charbm{}-5} &
   16.83 & 15.92 & 18.42 & 22.39 & 18.55 & \textbf{18.48} & \textcolor{red}{$-$49.2} &
   13.88 & 9.01 & 9.95 & 14.08 & 12.89 & \textbf{11.67} & \textcolor{red}{$-$29.8} \\
 & DN-3 &
   15.88 & 16.08 & 18.36 & 21.55 & 17.97 & 18.05 & \textcolor{red}{$-$45.7} &
   13.23 & 9.62 & 10.71 & 12.97 & 12.54 & 11.58 & \textcolor{red}{$-$28.9} \\
 & DN-5 &
   14.11 & 17.04 & 18.13 & 21.13 & 18.17 & 17.89 & \textcolor{red}{$-$44.4} &
   13.17 & 10.10 & 10.76 & 13.15 & 13.00 & 11.85 & \textcolor{red}{$-$31.8} \\
\addlinespace[2pt]
\multirow{7}{*}{Qwen3-4B}
 & ZS &
   13.82 & 12.81 & 14.44 & 16.61 & 13.32 & 14.19 & \textcolor{red}{$-$14.5} &
   13.84 & 9.67 & 9.48 & 11.59 & 11.79 & 10.98 & \textcolor{red}{$-$22.2} \\
 & RS-3 &
   13.49 & 12.09 & 14.12 & 17.42 & 12.63 & 13.94 & \textcolor{red}{$-$12.5} &
   12.31 & 8.10 & 8.71 & 11.43 & 11.16 & 10.06 & \textcolor{red}{$-$12.0} \\
 & RS-5 &
   13.31 & 11.81 & 14.23 & 17.01 & 12.37 & 13.73 & \textcolor{red}{$-$10.8} &
   12.18 & 8.17 & 8.90 & 11.72 & 11.50 & 10.24 & \textcolor{red}{$-$13.9} \\
 & \textit{\charbm{}-3} &
   10.92 & 11.11 & 13.47 & 16.02 & 11.83 & 12.71 & \textcolor{red}{$-$2.6} &
   10.81 & 8.07 & 8.61 & 10.37 & 10.95 & 9.58 & \textcolor{red}{$-$6.6} \\
 & \textit{\charbm{}-5} &
   10.77 & 11.28 & 13.27 & 16.50 & 11.83 & \textbf{12.79} & \textcolor{red}{$-$3.2} &
   10.61 & 7.89 & 8.45 & 10.54 & 9.96 & \textbf{9.32} & \textcolor{red}{$-$3.7} \\
 & DN-3 &
   10.84 & 11.43 & 13.24 & 15.78 & 11.95 & 12.70 & \textcolor{red}{$-$2.5} &
   10.80 & 8.33 & 8.36 & 10.95 & 10.89 & 9.72 & \textcolor{red}{$-$8.2} \\
 & DN-5 &
   10.42 & 11.74 & 13.43 & 16.16 & 12.36 & 12.91 & \textcolor{red}{$-$4.2} &
   10.38 & 8.98 & 8.46 & 10.81 & 10.25 & 9.70 & \textcolor{red}{$-$7.9} \\
\addlinespace[2pt]
\multirow{7}{*}{Gemma-3-4B}
 & ZS &
   16.88 & 15.38 & 18.59 & 19.30 & 15.93 & 17.17 & \textcolor{red}{$-$38.6} &
   19.91 & 15.15 & 17.34 & 17.65 & 18.63 & 17.39 & \textcolor{red}{$-$93.5} \\
 & RS-3 &
   15.78 & 14.47 & 18.63 & 19.64 & 16.84 & 17.06 & \textcolor{red}{$-$37.7} &
   19.08 & 14.35 & 14.45 & 18.52 & 17.66 & 16.51 & \textcolor{red}{$-$83.7} \\
 & RS-5 &
   16.11 & 15.05 & 19.04 & 20.79 & 17.20 & 17.64 & \textcolor{red}{$-$42.4} &
   19.67 & 14.49 & 15.10 & 20.07 & 18.73 & 17.30 & \textcolor{red}{$-$92.5} \\
 & \textit{\charbm{}-3} &
   13.99 & 13.67 & 18.15 & 18.65 & 15.72 & 16.05 & \textcolor{red}{$-$29.5} &
   17.85 & 13.75 & 13.80 & 18.70 & 16.81 & 15.95 & \textcolor{red}{$-$77.5} \\
 & \textit{\charbm{}-5} &
   12.98 & 13.73 & 18.21 & 19.05 & 15.27 & \textbf{15.90} & \textcolor{red}{$-$28.3} &
   18.20 & 14.07 & 13.95 & 17.74 & 17.21 & \textbf{15.97} & \textcolor{red}{$-$77.7} \\
 & DN-3 &
   12.30 & 15.19 & 17.10 & 19.31 & 15.93 & 16.12 & \textcolor{red}{$-$30.1} &
   15.58 & 13.89 & 14.41 & 16.36 & 15.73 & 15.10 & \textcolor{red}{$-$68.0} \\
 & DN-5 &
   12.16 & 15.30 & 16.54 & 18.25 & 15.58 & 15.72 & \textcolor{red}{$-$26.8} &
   15.57 & 14.55 & 14.62 & 17.29 & 16.91 & 15.75 & \textcolor{red}{$-$75.3} \\
\addlinespace[2pt]
\multirow{7}{*}{Llama-3.1-8B}
 & ZS &
   17.87 & 14.63 & 16.68 & 18.92 & 16.11 & 16.76 & \textcolor{red}{$-$35.2} &
   17.96 & 11.63 & 12.58 & 14.69 & 14.56 & 13.83 & \textcolor{red}{$-$53.9} \\
 & RS-3 &
   15.82 & 13.96 & 16.16 & 18.59 & 15.19 & 15.93 & \textcolor{red}{$-$28.5} &
   16.80 & 10.56 & 11.63 & 14.88 & 13.71 & 13.09 & \textcolor{red}{$-$45.7} \\
 & RS-5 &
   15.08 & 13.73 & 15.82 & 18.95 & 15.47 & 15.82 & \textcolor{red}{$-$27.7} &
   16.68 & 10.57 & 11.57 & 15.32 & 13.75 & 13.18 & \textcolor{red}{$-$46.6} \\
 & \textit{\charbm{}-3} &
   12.92 & 13.47 & 14.96 & 17.53 & 14.70 & 14.79 & \textcolor{red}{$-$19.4} &
   13.80 & 10.34 & 11.21 & 14.54 & 13.39 & 12.45 & \textcolor{red}{$-$38.5} \\
 & \textit{\charbm{}-5} &
   11.33 & 13.85 & 14.67 & 17.58 & 13.78 & \textbf{14.38} & \textcolor{red}{$-$16.0} &
   13.47 & 9.80 & 10.92 & 14.34 & 12.94 & \textbf{12.08} & \textcolor{red}{$-$34.4} \\
 & DN-3 &
   12.29 & 13.21 & 15.12 & 17.87 & 14.47 & 14.69 & \textcolor{red}{$-$18.5} &
   13.74 & 11.21 & 11.51 & 13.83 & 12.91 & 12.48 & \textcolor{red}{$-$38.9} \\
 & DN-5 &
   12.70 & 13.08 & 14.76 & 17.73 & 14.54 & 14.65 & \textcolor{red}{$-$18.2} &
   13.15 & 11.30 & 11.44 & 14.04 & 12.99 & 12.48 & \textcolor{red}{$-$38.9} \\
\addlinespace[2pt]
\multirow{7}{*}{Qwen3-8B}
 & ZS &
   13.29 & 11.78 & 14.48 & 15.65 & 12.82 & 13.57 & \textcolor{red}{$-$9.5} &
   13.68 & 9.90 & 9.56 & 12.03 & 11.74 & 11.12 & \textcolor{red}{$-$23.7} \\
 & RS-3 &
   12.80 & 11.63 & 14.21 & 15.68 & 12.16 & 13.28 & \textcolor{red}{$-$7.2} &
   13.62 & 8.71 & 9.36 & 11.81 & 11.92 & 10.75 & \textcolor{red}{$-$19.6} \\
 & RS-5 &
   12.81 & 11.70 & 13.89 & 15.83 & 12.78 & 13.39 & \textcolor{red}{$-$8.1} &
   13.87 & 9.27 & 9.13 & 12.02 & 11.96 & 10.93 & \textcolor{red}{$-$21.7} \\
 & \textit{\charbm{}-3} &
   10.48 & 11.05 & 13.37 & 14.70 & 11.67 & 12.30 & 0.7 &
   11.31 & 8.53 & 8.71 & 10.89 & 10.64 & \textbf{9.83} & \textcolor{red}{$-$9.4} \\
 & \textit{\charbm{}-5} &
   10.19 & 11.21 & 12.81 & 14.55 & 11.93 & \textbf{12.21} & 1.5 &
   11.16 & 8.28 & 8.59 & 11.43 & 10.79 & 9.87 & \textcolor{red}{$-$9.9} \\
 & DN-3 &
   10.74 & 11.29 & 13.11 & 15.05 & 11.85 & 12.46 & \textcolor{red}{$-$0.6} &
   11.41 & 8.99 & 9.22 & 10.81 & 11.11 & 10.14 & \textcolor{red}{$-$12.9} \\
 & DN-5 &
   10.18 & 11.17 & 13.03 & 14.67 & 11.98 & 12.28 & 0.9 &
   11.61 & 9.75 & 8.94 & 10.95 & 11.40 & 10.40 & \textcolor{red}{$-$15.7} \\
\addlinespace[2pt]
\multirow{7}{*}{Qwen3-14B}
 & ZS &
   11.71 & 10.25 & 11.93 & 13.79 & 11.11 & 11.74 & 5.3 &
   13.18 & 8.50 & 9.57 & 10.56 & 11.31 & 10.29 & \textcolor{red}{$-$14.5} \\
 & RS-3 &
   11.01 & 9.77 & 11.51 & 13.56 & 10.56 & 11.28 & 9.0 &
   12.77 & 8.37 & 8.94 & 10.68 & 11.11 & 10.06 & \textcolor{red}{$-$12.0} \\
 & RS-5 &
   10.75 & 9.60 & 11.29 & 13.30 & 10.68 & 11.12 & 10.2 &
   12.76 & 8.36 & 8.82 & 10.59 & 10.90 & 9.97 & \textcolor{red}{$-$11.0} \\
 & \textit{\charbm{}-3} &
   8.42 & 9.15 & 10.19 & 12.37 & 9.72 & 10.04 & 19.0 &
   10.62 & 7.75 & 8.18 & 9.59 & 10.09 & 9.05 & \textcolor{red}{$-$0.7} \\
 & \textit{\charbm{}-5} &
   7.87 & 8.91 & 9.89 & 12.05 & 9.56 & \textbf{9.73} & 21.4 &
   10.35 & 7.70 & 7.97 & 9.51 & 10.02 & \textbf{8.93} & 0.7 \\
 & DN-3 &
   8.67 & 9.44 & 10.74 & 12.97 & 9.91 & 10.41 & 16.0 &
   10.93 & 8.22 & 8.56 & 9.83 & 10.07 & 9.33 & \textcolor{red}{$-$3.8} \\
 & DN-5 &
   8.11 & 9.30 & 10.52 & 12.58 & 9.85 & 10.15 & 18.1 &
   10.62 & 8.16 & 8.39 & 9.74 & 9.98 & 9.21 & \textcolor{red}{$-$2.5} \\
\addlinespace[2pt]
\multirow{7}{*}{Gemma-3-12B}
 & ZS &
   11.60 & 9.81 & 12.37 & 13.22 & 10.41 & 11.43 & 7.8 &
   16.53 & 10.95 & 12.34 & 12.97 & 13.48 & 12.85 & \textcolor{red}{$-$42.9} \\
 & RS-3 &
   10.77 & 9.11 & 11.75 & 12.90 & 10.08 & 10.88 & 12.2 &
   16.63 & 10.33 & 11.45 & 12.30 & 11.49 & 11.98 & \textcolor{red}{$-$33.3} \\
 & RS-5 &
   10.51 & 8.93 & 11.30 & 12.80 & 9.83 & 10.65 & 14.1 &
   16.06 & 10.25 & 11.40 & 12.40 & 12.05 & 12.01 & \textcolor{red}{$-$33.6} \\
 & \textit{\charbm{}-3} &
   7.89 & 8.36 & 10.37 & 12.19 & 9.15 & 9.65 & 22.1 &
   12.48 & 9.40 & 10.37 & 11.13 & 11.13 & 10.69 & \textcolor{red}{$-$19.0} \\
 & \textit{\charbm{}-5} &
   7.66 & 8.36 & 9.76 & 11.48 & 8.82 & \textbf{9.27} & 25.2 &
   12.21 & 9.42 & 10.31 & 11.03 & 10.57 & \textbf{10.52} & \textcolor{red}{$-$17.0} \\
 & DN-3 &
   7.93 & 8.72 & 10.78 & 12.28 & 9.04 & 9.80 & 20.9 &
   12.55 & 9.93 & 10.94 & 11.19 & 10.68 & 10.87 & \textcolor{red}{$-$21.0} \\
 & DN-5 &
   7.67 & 8.64 & 10.85 & 11.73 & 9.48 & 9.73 & 21.5 &
   12.08 & 9.88 & 10.50 & 11.00 & 10.95 & 10.73 & \textcolor{red}{$-$19.4} \\
\addlinespace[2pt]
\multirow{7}{*}{Sarvam-m}
 & ZS &
   13.33 & 13.96 & 14.35 & 15.48 & 13.49 & \textbf{14.16} & \textcolor{red}{$-$14.2} &
   14.89 & 10.25 & 10.56 & 11.64 & 12.17 & 11.56 & \textcolor{red}{$-$28.7} \\
 & RS-3 &
   15.54 & 14.00 & 16.08 & 18.71 & 14.85 & 15.83 & \textcolor{red}{$-$27.7} &
   15.69 & 9.63 & 10.24 & 12.21 & 12.33 & 11.58 & \textcolor{red}{$-$28.9} \\
 & RS-5 &
   15.65 & 14.43 & 16.01 & 18.79 & 14.87 & 15.95 & \textcolor{red}{$-$28.7} &
   16.13 & 9.89 & 10.16 & 12.49 & 11.70 & 11.63 & \textcolor{red}{$-$29.4} \\
 & \textit{\charbm{}-3} &
   13.40 & 13.96 & 15.47 & 17.55 & 13.65 & 14.85 & \textcolor{red}{$-$19.8} &
   13.48 & 9.57 & 9.72 & 11.83 & 12.03 & 11.05 & \textcolor{red}{$-$23.0} \\
 & \textit{\charbm{}-5} &
   12.62 & 13.69 & 14.79 & 17.72 & 13.89 & 14.62 & \textcolor{red}{$-$18.0} &
   13.16 & 9.41 & 9.52 & 12.12 & 11.83 & \textbf{10.96} & \textcolor{red}{$-$21.9} \\
 & DN-3 &
   12.38 & 14.88 & 15.95 & 17.83 & 14.29 & 15.18 & \textcolor{red}{$-$22.5} &
   13.47 & 10.49 & 10.33 & 11.59 & 11.76 & 11.31 & \textcolor{red}{$-$25.9} \\
 & DN-5 &
   11.33 & 14.93 & 15.47 & 17.84 & 13.92 & 14.86 & \textcolor{red}{$-$19.9} &
   13.04 & 10.35 & 10.48 & 11.52 & 11.84 & 11.25 & \textcolor{red}{$-$25.2} \\
\addlinespace[2pt]
\multirow{7}{*}{Qwen3-32B}
 & ZS &
   13.64 & 11.75 & 13.65 & 16.10 & 12.50 & 13.50 & \textcolor{red}{$-$9.0} &
   15.85 & 9.63 & 10.44 & 11.77 & 12.06 & 11.51 & \textcolor{red}{$-$28.1} \\
 & RS-3 &
   12.58 & 10.91 & 12.47 & 15.34 & 11.83 & 12.61 & \textcolor{red}{$-$1.8} &
   15.17 & 8.90 & 10.18 & 12.12 & 11.89 & 11.21 & \textcolor{red}{$-$24.7} \\
 & RS-5 &
   12.00 & 10.73 & 12.05 & 15.08 & 11.42 & 12.26 & 1.1 &
   14.93 & 8.82 & 9.95 & 11.63 & 11.68 & 10.97 & \textcolor{red}{$-$22.1} \\
 & \textit{\charbm{}-3} &
   8.64 & 9.87 & 10.86 & 13.77 & 10.49 & 10.82 & 12.7 &
   11.41 & 8.24 & 8.87 & 10.30 & 10.95 & 9.74 & \textcolor{red}{$-$8.3} \\
 & \textit{\charbm{}-5} &
   7.97 & 9.25 & 10.33 & 13.09 & 10.11 & \textbf{10.25} & 17.3 &
   11.29 & 7.82 & 8.90 & 9.84 & 10.76 & \textbf{9.48} & \textcolor{red}{$-$5.5} \\
 & DN-3 &
   9.33 & 10.22 & 11.40 & 14.42 & 10.77 & 11.31 & 8.7 &
   12.16 & 8.62 & 9.49 & 10.58 & 10.92 & 10.10 & \textcolor{red}{$-$12.4} \\
 & DN-5 &
   8.42 & 10.02 & 11.08 & 14.29 & 10.77 & 11.03 & 11.0 &
   11.42 & 8.46 & 9.15 & 9.98 & 10.66 & 9.72 & \textcolor{red}{$-$8.2} \\
\addlinespace[2pt]
\multirow{7}{*}{\textbf{Gemma-3-27B}}
 & ZS &
   9.94 & 8.80 & 10.86 & 11.58 & 9.35 & 10.08 & 18.7 &
   15.92 & 9.96 & 11.73 & 11.88 & 12.10 & 11.88 & \textcolor{red}{$-$32.2} \\
 & RS-3 &
   8.59 & 8.04 & 9.44 & 10.77 & 8.37 & 9.04 & 27.0 &
   13.88 & 9.24 & 10.54 & 10.36 & 10.29 & 10.52 & \textcolor{red}{$-$17.0} \\
 & RS-5 &
   8.18 & 7.71 & 9.19 & 10.38 & 8.04 & 8.70 & 29.8 &
   13.31 & 8.93 & 9.83 & 10.13 & 9.99 & 10.11 & \textcolor{red}{$-$12.5} \\
 & \textit{\charbm{}-3} &
   6.10 & 7.24 & 8.48 & 9.66 & 7.45 & 7.85 & 36.7 &
   10.35 & 8.99 & 9.14 & 9.37 & 9.64 & 9.40 & \textcolor{red}{$-$4.6} \\
 & \textit{\charbm{}-5} &
   5.55 & 6.85 & 7.78 & 8.98 & 7.37 & \textbf{7.38} & 40.4 &
   9.90 & 8.33 & 8.73 & 8.92 & 9.21 & \textbf{8.90} & 0.9 \\
 & DN-3 &
   6.34 & 7.71 & 8.74 & 10.07 & 7.85 & 8.21 & 33.7 &
   11.30 & 8.86 & 9.48 & 9.21 & 9.63 & 9.51 & \textcolor{red}{$-$5.8} \\
 & DN-5 &
   5.57 & 7.49 & 8.35 & 9.75 & 7.40 & 7.80 & 37.0 &
   10.53 & 8.63 & 9.03 & 8.79 & 9.76 & 9.20 & \textcolor{red}{$-$2.3} \\
\addlinespace[2pt]
\bottomrule
\end{tabular}}
\end{table*}

\paragraph{Finding 1: Model scale is the dominant factor.}
Within the Gemma-3 family, WER decreases monotonically with scale
across every retrieval setting and both languages: 4B $>$ 12B $>$ 27B.
Under \charbm{}-5 ($n{=}3$), Gemma-3-27B achieves Hindi WER of 16.22\%
($\Delta{=}55.0\%$) and Marathi WER of 23.42\% ($\Delta{=}33.3\%$),
versus Gemma-3-12B at 19.69\%/26.69\% and Gemma-3-4B at
28.05\%/34.76\%.
The Qwen3 family confirms the same trend: Hindi \charbm{}-5 WER decreases
from 27.68\% (4B) $\to$ 25.26\% (8B) $\to$ 20.65\% (14B) $\to$
19.93\% (32B).
Scale effects are larger on Hindi than Marathi, consistent with
greater Hindi pretraining coverage across all model families.

\begin{figure}[t]
  \centering
  \includegraphics[width=0.85\linewidth]{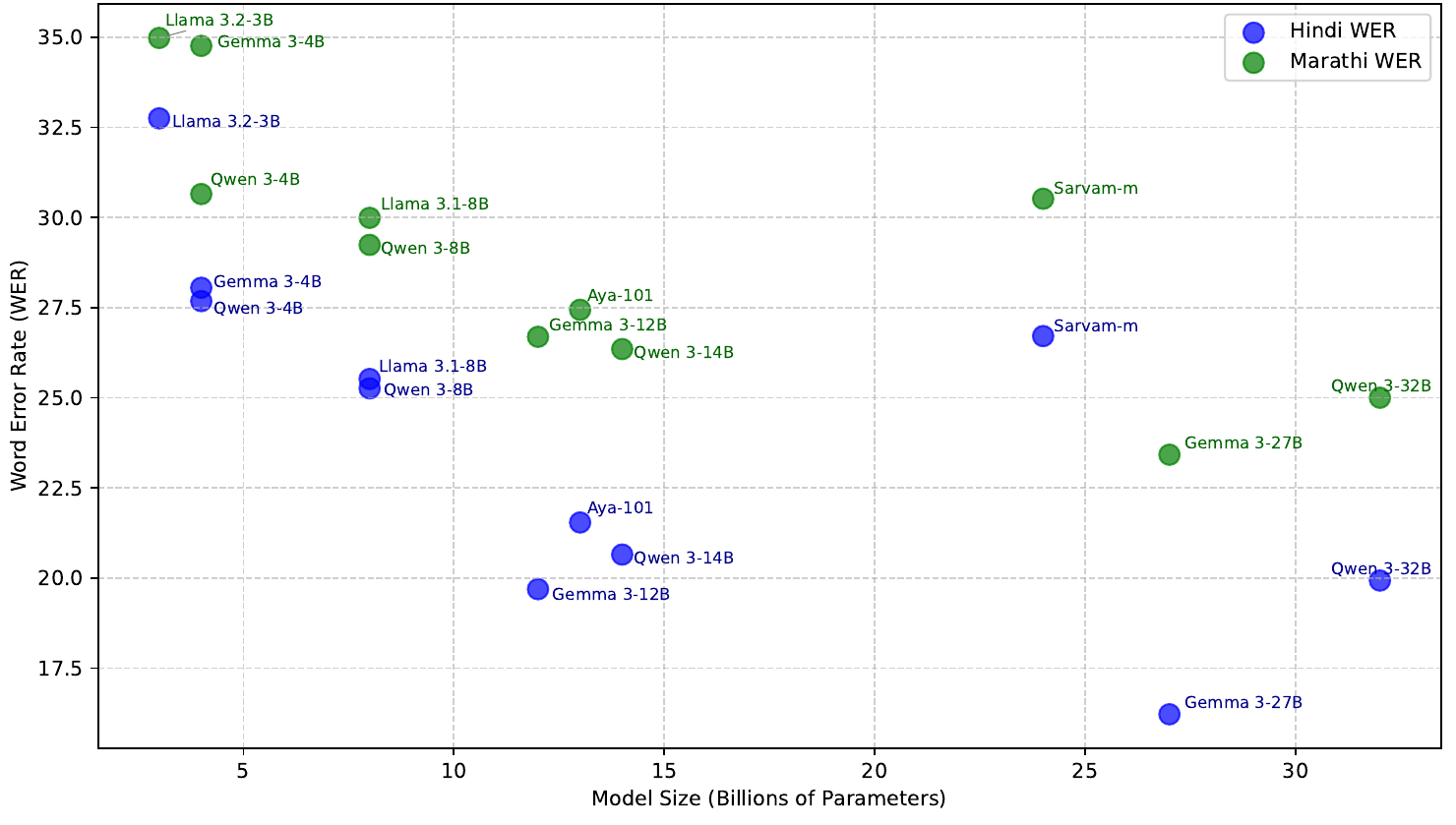}
  \caption{WER (\%) under \charbm{}-5 ($n{=}3$) vs.\ model
  parameter count (log scale).
  Both languages follow a log-linear decrease.}
  \label{fig:scale_wer}
\end{figure}

\paragraph{Finding 2: General-purpose scale outperforms
Indic specialization.}
Sarvam-m (24B, Indic-specialized) achieves Hindi \charbm{}-5 WER of
26.71\% and Marathi 30.52\%, substantially worse than Gemma-3-27B
(16.22\%/23.42\%) and Qwen3-32B (19.93\%/25.00\%), neither of
which has Indic-specific training.
Even Qwen3-14B (14B) outperforms Sarvam-m on Hindi
(20.65\% vs.\ 26.71\%).
Instruction-following capacity and broad multilingual pretraining
appear more important than script-specific pretraining in the
prompted, zero-fine-tuning regime.

\paragraph{Finding 3: Few-shot gains are capacity dependent.}
Models with ${\geq}$8B parameters show consistent WER improvement
from ZS to \charbm{}-5: Qwen3-8B $\Delta_\text{WER}$ increases from
15.9\% (ZS) to 30.0\% (\charbm{}-5) on Hindi.
For models below 8B, Marathi frequently degrades below the OCR
baseline: Llama-3.2-3B ZS yields $\Delta_\text{WER}{=}{-11.4\%}$
on Marathi, with \charbm{}-5 recovering only to $+0.4\%$.
Gemma-3-4B shows the same pattern (ZS: $-9.7\%$, \charbm{}-5: $+1.0\%$).
This threshold is consistent across the Qwen3 and Gemma-3 families
and points to a minimum capacity requirement for reliable
Devanagari few-shot correction.

\begin{figure}[t]
  \centering
  \includegraphics[width=\linewidth]{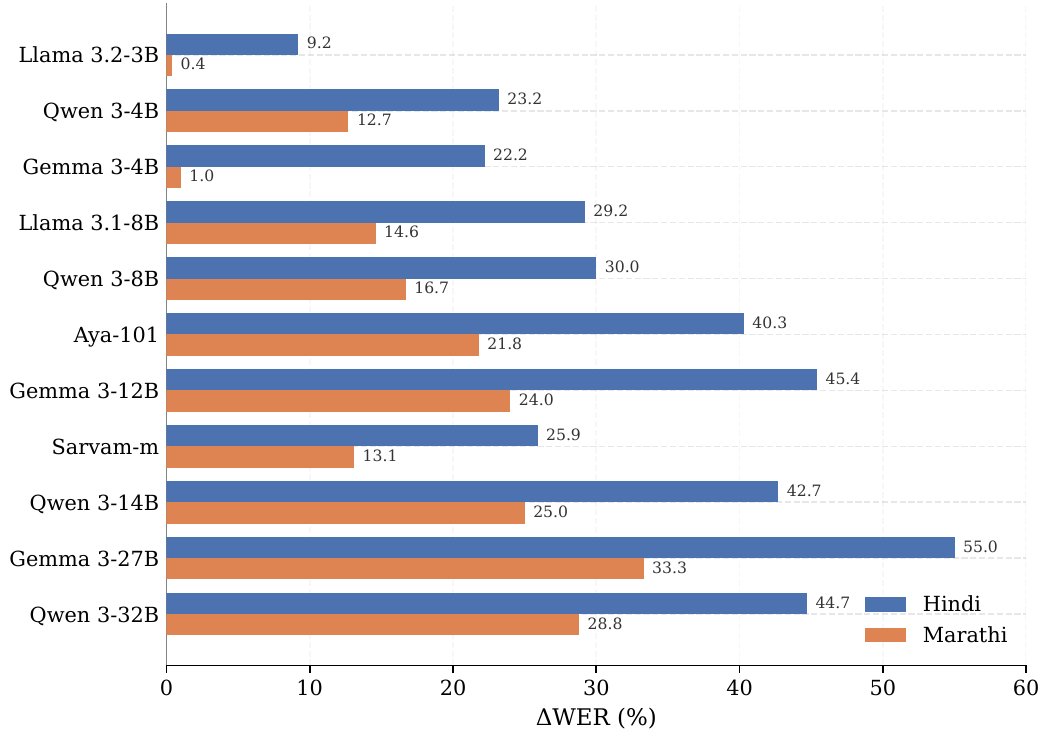}
  \caption{$\Delta_\text{WER}$ (\%) for \charbm{}-5
  ($n{=}3$) across all 11 models, ordered by parameter count.}
  \label{fig:delta_wer_grouped}
\end{figure}

\paragraph{Finding 4: \charbm{} consistently outperforms
random selection.}
Across all ${\geq}$8B models, \charbm{}-5 outperforms RS-5 by
2.8--4.0\,pp absolute WER on Hindi and 2.9--3.8\,pp on Marathi.
The gap is largest for Qwen3-32B (\charbm{}-5: 19.93\% vs.\ RS-5:
23.94\% Hindi, a 4.01\,pp gap) and smallest for Gemma-3-27B (\charbm{}-5: 16.22\% vs.\ RS-5: 18.99\%, a 2.77\,pp gap), where the larger model's stronger zero-shot prior leaves less headroom for retrieval to exploit.
For sub-8B models, the gap is less consistent, but \charbm{}-5 still outperforms RS-5
in the majority of cases, consistent with the capacity threshold
in Finding~3.

\paragraph{Finding 5: Marathi is consistently harder
than Hindi.}
$\Delta_\text{WER}$ on Marathi trails Hindi by 7--22\,pp across
all models and settings.
The gap grows with scale: Llama-3.2-3B shows an 8.8\,pp gap
(9.2\% Hindi vs.\ 0.4\% Marathi at \charbm{}-5); at Gemma-3-27B it
widens to 21.7\,pp (55.0\% vs.\ 33.3\%), as larger models benefit Hindi disproportionately more than Marathi. We attribute this to Marathi's greater morphological complexity:
agglutinative verb forms and Marathi-specific conjunct clusters
produce longer and more varied error spans, requiring stronger
generalization from retrieved examples.

\begin{figure}[t]
  \centering
  \includegraphics[width=\linewidth]{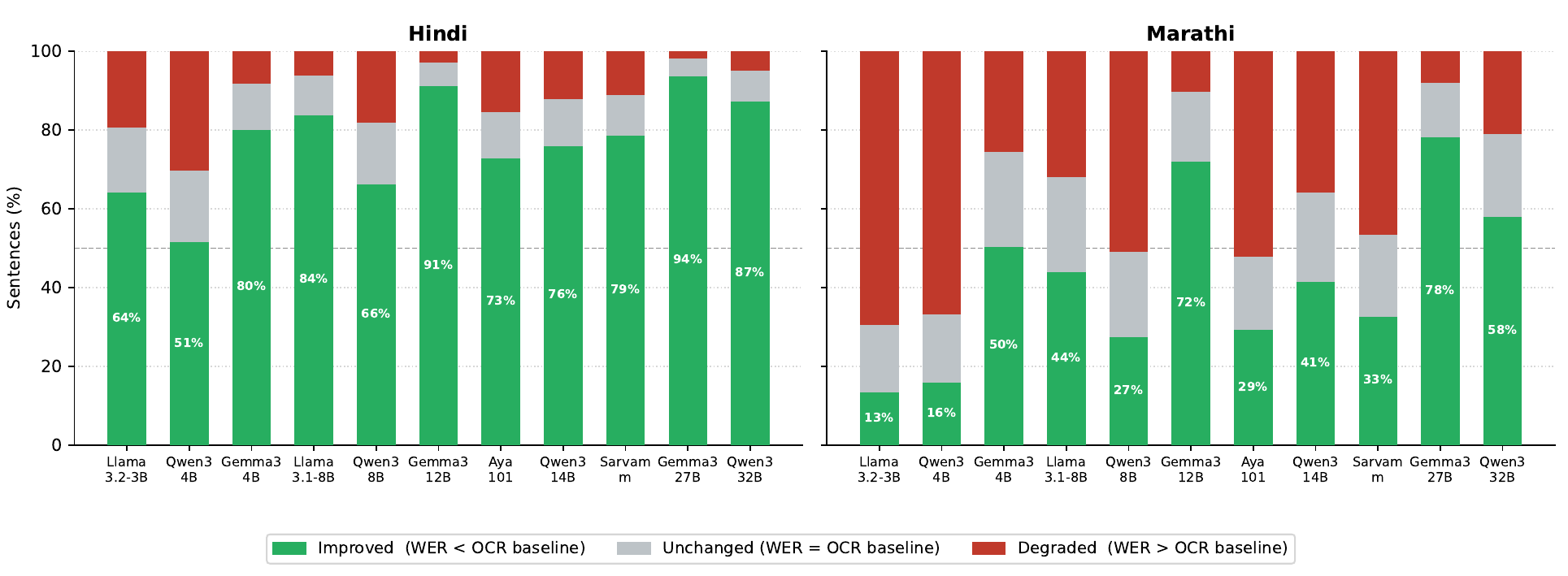}
  \caption{Per-sentence outcome distribution under \charbm{}-5
  ($n{=}3$, $k{=}5$). green = sentences where model WER is
  below the OCR baseline; grey = unchanged; red = degraded.}
  \label{fig:sentence_outcomes}
\end{figure}

\paragraph{Per-sentence benefit analysis.}
Figure~\ref{fig:sentence_outcomes} reports the fraction of
sentences improved, unchanged, or degraded under \charbm{}-5
($n{=}3$, $k{=}5$) relative to the OCR baseline.
The capacity threshold is immediately visible: smaller models
degrade a substantial fraction of Marathi sentences, with
degradation rates exceeding improvement rates for
Llama-3.2-3B and Qwen3-4B.
For models with 12B or more parameters, the Hindi improvement
rate reaches 76--94\%, though Marathi improvement rates vary
more widely (42--78\%), reflecting the persistent difficulty of Marathi correction identified in Finding~5. This confirms that above this threshold, \charbm{} retrieval provides consistent sentence-level gains rather than aggregate improvements driven by a subset of easy sentences.
Sarvam-m is the notable exception among large models:
despite its 24B scale it achieves improvement rates closer to
the 8B tier than to Gemma-3-27B, consistent with its weaker
aggregate WER in Finding~2.

\subsection{Retrieval and N-gram Ablation}
\label{sec:results:ablation}

\paragraph{N-gram order.}
$n{=}3$ achieves the lowest WER in 9 of 11 models at $k{=}5$ on
Hindi (avg.\ 24.00\% vs.\ 24.42\% for $n{=}2$ and 25.26\% for
$n{=}1$) and in all 11 models on Marathi (avg.\ 29.01\% vs.\
29.45\% for $n{=}2$).
The $n{=}1{\to}2$ gain is consistently larger than $n{=}2{\to}3$,
reflecting diminishing returns: unigrams are non-discriminative
(common matras appear everywhere), bigrams capture
consonant--matra adjacency, and trigrams add the
consonant--matra--consonant neighborhood that fully encodes
whether a matra is present, absent, or substituted.
Marathi benefits more from a higher $n$ (avg.\ $n{=}2{\to}3$ gain:
0.44\,pp Hindi vs.\ 0.72\,pp Marathi at $k{=}5$), consistent
with Marathi's more complex conjunct environment requiring
broader character context for discrimination.

\begin{figure}[t]
  \centering
  \includegraphics[width=\linewidth]{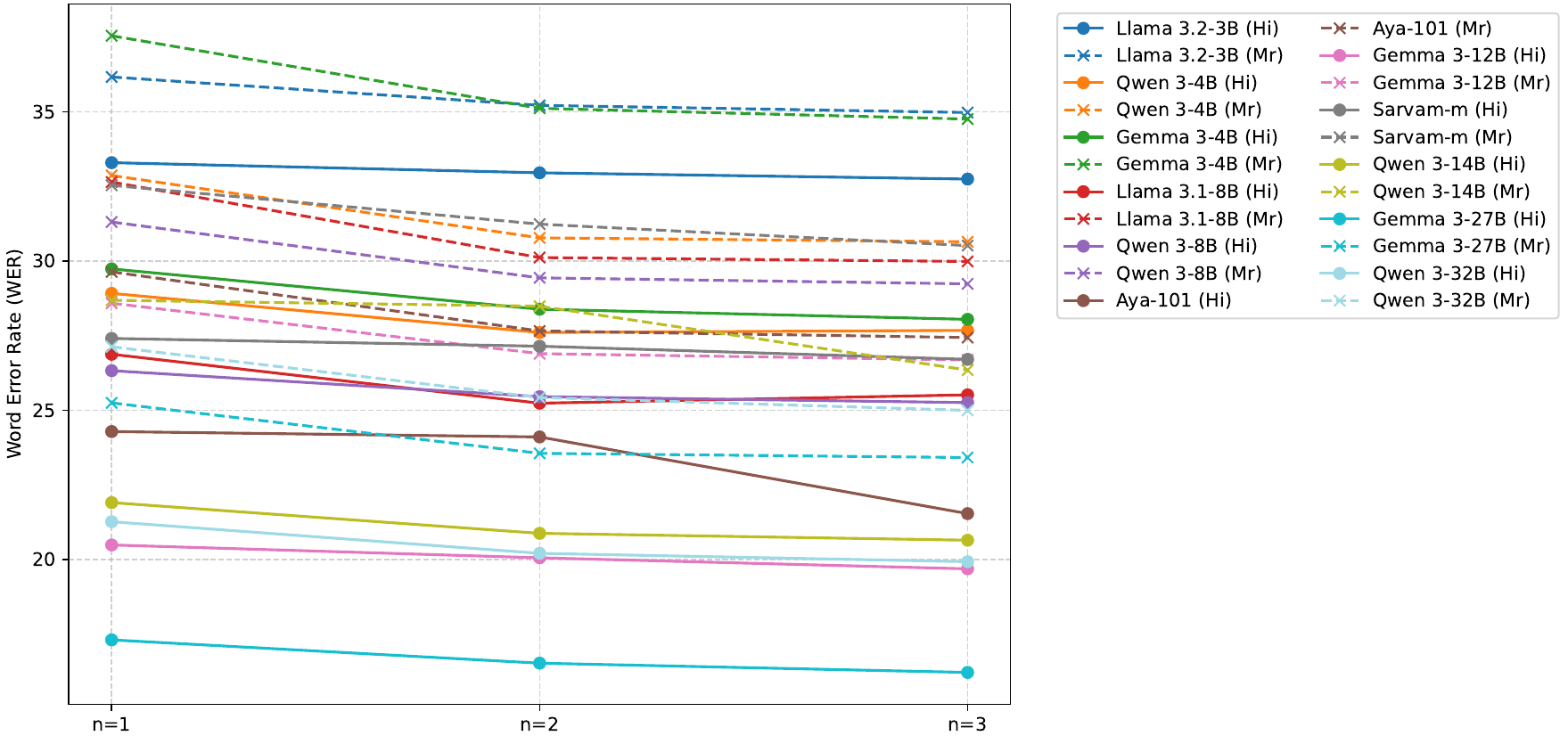}
  \caption{WER (\%) under \charbm{}-5 as $n$-gram order increases
  from 1 to 3 (left: Hindi, right: Marathi).
  Each line is one model; bold lines are Gemma-3-27B and
  Qwen3-32B.
  Dashed horizontal line = OCR baseline.
  $n{=}3$ dominates across all models; Marathi shows larger
  absolute gains from higher $n$.}
  \label{fig:ngram_slope}
\end{figure}

\paragraph{Shot count.}
$k{=}5$ outperforms $k{=}3$ by 0.3--2.0\,pp WER at $n{=}3$
across all models and languages, confirming that additional
retrieved examples provide complementary corrective signal.
The $k{=}3{\to}5$ gain is larger for \charbm{} than for random
selection (avg.\ 0.76\,pp vs.\ 0.42\,pp on Hindi), suggesting
BM25 retrieval returns meaningfully diverse examples at higher
$k$ rather than near-duplicates.

\paragraph{\charbm{} vs.\ dense retrieval.}
At $k{=}5$, \charbm{}-5 ($n{=}3$) matches or outperforms dense retrieval on Hindi for 10 of 11 models. On Marathi, \charbm{}-5 likewise matches or outperforms dense retrieval for 10 of 11 models; the sole exception is Gemma-3-4B, where dense retrieval wins by 0.05\,pp. The practical advantage of \charbm{} remains decisive: index construction completes in under 1\,s and per-query retrieval in under 1\,ms, versus 30--60\,s and 5--15\,ms for dense retrieval, with no GPU required.

\subsection{Error Analysis}
\label{sec:results:error}

\subsubsection{Quantitative Error Analysis}
\label{sec:results:error:quant}

\begin{figure}[t]
  \centering
  \includegraphics[width=0.85\linewidth]{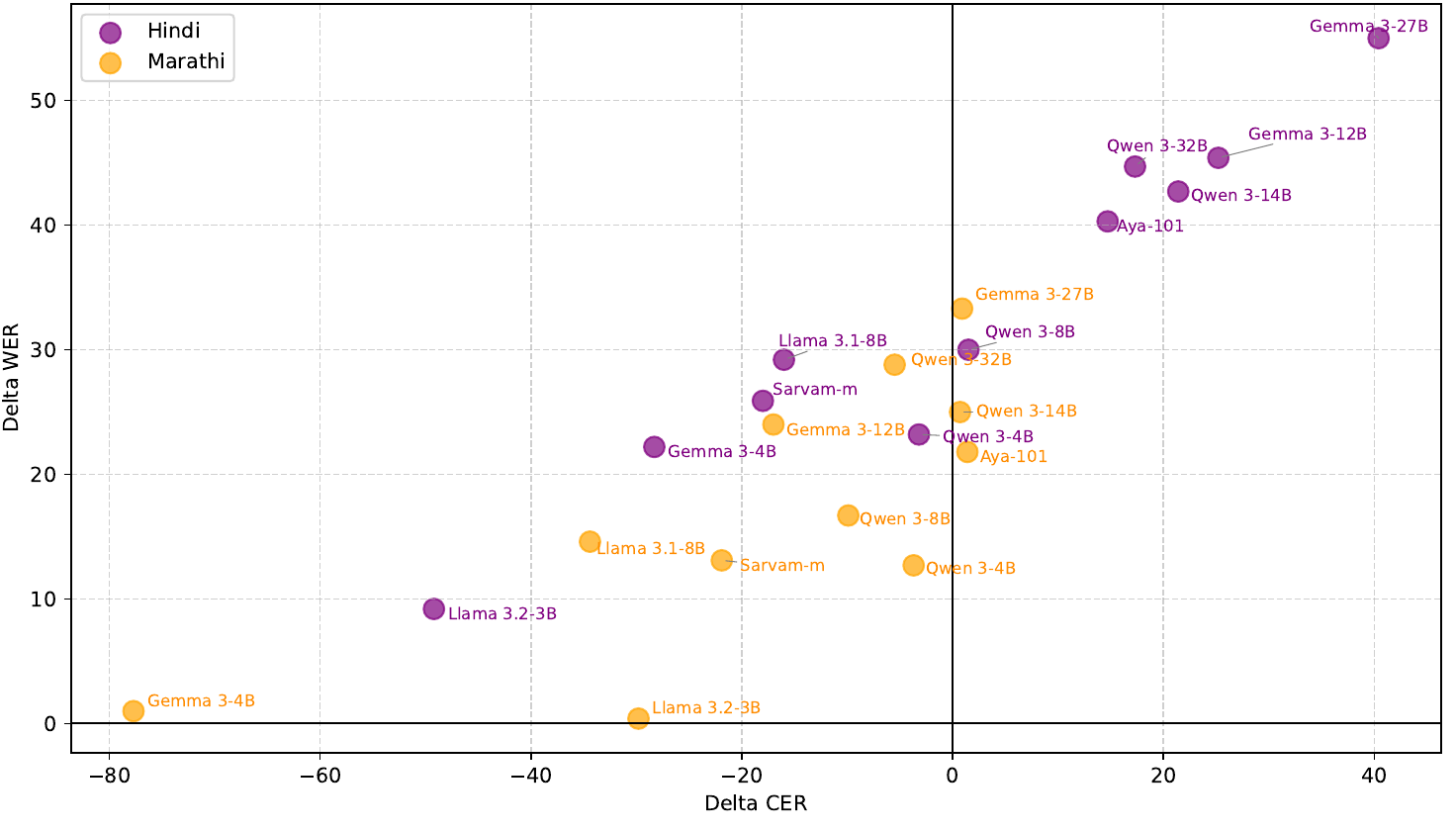}
  \caption{$\Delta_\text{WER}$ vs.\ $\Delta_\text{CER}$ under
  \charbm{}-5 ($n{=}3$) for all 11 models (Hindi: circles,
  Marathi: triangles).
  Points below the $y{=}x$ diagonal indicate models where
  WER improvement exceeds CER improvement---the hallucination
  regime where word-level fluency is restored but character-exact
  accuracy is not.
  Sub-8B models cluster below the diagonal; large models
  approach it.}
  \label{fig:wer_cer_scatter}
\end{figure}

\paragraph{WER/CER analysis.}
A persistent asymmetry between WER and CER improvements is
visible across all models in Table~\ref{tab:cer_hin_mar}.
Large WER reductions frequently co-occur with smaller or
negative CER gains, indicating that word-level fluency is
recovered before character-exact restoration.
This is most pronounced for sub-8B models: Llama-3.2-3B under
\charbm{}-5 achieves $\Delta_\text{WER}{=}9.2\%$ on Hindi but CER
\emph{increases} by 49.2\% relative to baseline, meaning the
model replaces OCR-corrupted tokens with fluent but
orthographically incorrect substitutions.
For Gemma-3-27B \charbm{}-5, the gap closes substantially:
$\Delta_\text{WER}{=}55.0\%$ vs.\ $\Delta_\text{CER}{=}40.4\%$,
suggesting larger models achieve more faithful character-level
restoration alongside fluency correction.
Figure~\ref{fig:wer_cer_scatter} visualises this tension across
all 11 models and both languages: sub-8B models cluster below
the $y{=}x$ diagonal, while large models approach it.

\paragraph{Domain difficulty.}
D4 (Politics) is the hardest domain in both languages across all
models, driven by named-entity density.
Under Gemma-3-27B \charbm{}-5 on Hindi, D4 WER is 18.85\% vs.\
13.12\% for D1 (Astrology)---a 5.73\,pp gap that persists
across all retrieval strategies.
D1 (Astrology) and D5 (Sports) benefit most from \charbm{}
retrieval: formulaic sentence structures in horoscope
predictions and match reports produce high character-level
similarity between test sentences and the shot bank, making
BM25 retrieval particularly effective in these domains.
Named entity errors are the hardest category within D4: proper
nouns carry no corrective signal in character-level retrieval,
and the model must rely on world knowledge alone to restore
them, which fails systematically at all scales as evidenced
by the qualitative contrast in Table~\ref{tab:qual_lang}.

\subsubsection{Qualitative Analysis}
\label{sec:results:error:qual}

\paragraph{Scale effect on correction quality.}
Table~\ref{tab:qual_scale} shows the same Hindi sentence
processed by three Qwen3 models of increasing scale under
\charbm{}-5 ($n{=}3$, $k{=}5$).
The 4B model corrects none of the four OCR errors; the 14B
model resolves two (character confusion in \texthindi{जवाब}
and the verb form \texthindi{दिया}) while leaving the initial
character and plural form wrong; the 32B model achieves
perfect correction. This monotonic pattern holds across
the Gemma-3 family as well (Finding~1) and reflects the
minimum instruction-following capacity required to parse and
apply the corrective signal from retrieved examples.
\begin{table}[t]
\centering
\small
\caption{Scale effect: same Hindi OCR input under \charbm{}-5 ($n{=}3$, $k{=}5$) across Qwen3 models (Entertainment domain).}

\label{tab:qual_scale}
\begin{tabular}{@{}p{3cm}p{4cm}@{}}
\toprule
 & \textbf{Text} \\
\midrule
GT  & \texthindi{अब उन्होंने ट्रोल्स को जवाब दिया।}  \\
OCR & \texthindi{\textcolor{red}{अंब उन्हॉने} \textcolor{red}{ट्रोल्य} को
     \textcolor{red}{नवाब} \textcolor{red}{ट्रिया}।}  \\
\hdashline
Qwen3-4B  & \texthindi{\textcolor{red}{अंब} उन्होंने \textcolor{red}{ट्रोल} को
            \textcolor{red}{नवाब} \textcolor{red}{ट्रिया}।}
            \\
Qwen3-14B & \texthindi{\textcolor{red}{अंब} उन्होंने \textcolor{red}{ट्रोल} को
            \textcolor{green}{जवाब दिया}।}
           \\
Qwen3-32B & \texthindi{\textcolor{green}{अब} उन्होंने
            \textcolor{green}{ट्रोल्स} को
            \textcolor{green}{जवाब}
            \textcolor{green}{दिया}।}
           \\
\bottomrule
\end{tabular}
\end{table}

\paragraph{Retrieval strategy effect.}
Table~\ref{tab:qual_retrieval} isolates the contribution of
retrieval strategy by holding the model (Gemma-3-27B) and
shot count ($k{=}5$) fixed across all four settings on the
same Marathi sentence.
ZS completely hallucinates an unrelated sentence about water
supply disruption in Nagpur --- a failure mode consistent with
the domain difficulty finding for D4 (Politics).
RS-5 and \charbm{}-5 both achieve perfect correction, confirming that
any character-level context is sufficient for this error type
when the model is large enough.
DN-5 partially succeeds but substitutes \texthindi{खटके उडणार}
for the correct \texthindi{वाद होणार}, illustrating that
semantic proximity can introduce plausible but incorrect
alternatives when the error is orthographic rather than semantic.
\begin{table}[t]
\centering
\small
\caption{Retrieval strategy effect on a Marathi sentence
(Gemma-3-27B, $k{=}5$, Politics domain).}
\label{tab:qual_retrieval}
\begin{tabular}{@{}lp{6.5cm}r@{}}
\toprule
\textbf{Setting} & \textbf{Output} & \textbf{WER} \\
\midrule
GT  & \texthindi{एनडीएमध्ये जागावाटपावरून वाद होणार?}
    & 0.0 \\
OCR & \texthindi{एनडीएमध्ये \textcolor{red}{नागावाटपावरून}
     \textcolor{red}{बाढू} होणार?}
    & 50.0 \\
\hdashline
ZS        & \texthindi{\textcolor{red}{एनडीएमध्ये नागपूर शहराला
            पाणीपुरवठा खंडित होणार?}}
          & 100.0 \\
RS-5      & \texthindi{एनडीएमध्ये
            \textcolor{green}{जागावाटपावरून}
            \textcolor{green}{वाद} होणार?}
          & 0.0 \\
DN-5      & \texthindi{एनडीएमध्ये
            \textcolor{green}{जागावाटपावरून}
            \textcolor{red}{खटके उडणार?}}
          & 25.0 \\
\charbm{}-5 & \texthindi{एनडीएमध्ये
            \textcolor{green}{जागावाटपावरून}
            \textcolor{green}{वाद} होणार?}
          & 0.0 \\
\bottomrule
\end{tabular}
\end{table}
\paragraph{Hindi vs.\ Marathi contrast.}
Table~\ref{tab:qual_lang} pairs a Hindi Sports sentence and a
Marathi Sports sentence of comparable initial WER under
Gemma-3-27B \charbm{}-5.
For Hindi, seven of the eight OCR errors are character-level
distortions (matra substitution, glyph confusion) that
\charbm{} retrieval directly targets, and the model corrects
all but one, achieving WER 5.0\%.
For Marathi, the dominant errors are named entities
(\texthindi{रिक}$\to$\texthindi{रिंकू},
\texthindi{सरोन}$\to$\texthindi{सरोज}) where
character-level retrieval provides no corrective signal.
The model not only fails to restore the named entities but
introduces a new error (\texthindi{निक} for \texthindi{रिंकू}
and \texthindi{खास विवाह} for \texthindi{खासदार}), pushing
WER from 30.0\% to 40.0\%.
This degradation pattern is the primary driver of the
Hindi--Marathi gap quantified in Finding~5.
\begin{table}[t]
\centering
\small
\caption{Cross-language correction contrast,
Gemma-3-27B \charbm{}-5 ($n{=}3$, $k{=}5$).
Hindi corrects well; Marathi degrades due to named-entity
errors and morphological complexity.}
\label{tab:qual_lang}
\begin{tabular}{@{}p{1.5cm}p{1.5cm}p{9cm}@{}}
\toprule
\textbf{Lang} & & \textbf{Text}  \\
\midrule
\multirow{3}{*}{Hindi}
  & OCR
  & \texthindi{\textcolor{red}{द्रोनों} टीमें अपनी
    \textcolor{red}{तेंगारियों} में कोई कसर
    \textcolor{red}{नही'} छोड़ रही \textcolor{red}{हँ}
    \textcolor{red}{ऑर} प्रशेयकों को एक शानदार
    मुकाबले की उम्मीद \textcolor{red}{हे!}}
   \\
  & Pred
  & \texthindi{\textcolor{Forestgreen}{दोनों} टीमें
    अपनी \textcolor{Forestgreen}{तैयारियों} में कोई
    कसर नहीं छोड़ रही हैं
    \textcolor{Forestgreen}{और} प्रशंसकों को एक
    शानदार मुकाबले की उम्मीद है!}
  \\
  & GT
  & \texthindi{दोनों टीमें अपनी तैयारियों में कोई
    कसर नहीं छोड़ रही हैं और प्रशंसकों को एक
    शानदार मुकाबले की उम्मीद है।}
  \\
\midrule
\multirow{3}{*}{Marathi}
  & OCR
  & \texthindi{\textcolor{red}{रिक} सिंहचा खासद्वार
    \textcolor{red}{प्रिया सरोन} हिच्याबरोबर
    साखरपुडा सोहळा संपन्न झाला.}
  \\
  & Pred
  & \texthindi{\textcolor{red}{निक} सिंहचा खास विवाह
    \textcolor{red}{प्रिया सरोन} हिच्याबरोबर
    साखरपुडा सोहळा संपन्न झाला.}
  \\
  & GT
  & \texthindi{\textcolor{Forestgreen}{रिंकू} सिंहचा
    खासदार \textcolor{Forestgreen}{प्रिया सरोज}
    हिच्याबरोबर साखरपुडा सोहळा संपन्न झाला.}
  \\
\bottomrule
\end{tabular}
\end{table}

\section{Conclusion}
\label{sec:conclusion}

We evaluated open-weight LLMs for Devanagari
post-OCR correction in Hindi and Marathi, and introduced
\charbm{}, a character $n$-gram BM25 algorithm for few-shot
example selection.
Five findings emerge across 3B--32B models, seven retrieval
settings, and five domains.
Scale dominates: Gemma-3-27B under \charbm{}-5 ($n{=}3$)
reduces Hindi WER by 55.0\% and Marathi WER by 33.3\%.
General-purpose scale outperforms Indic specialisation:
Sarvam-m (24B) is substantially outperformed by both
Gemma-3-27B and Qwen3-32B.
Few-shot gains are capacity-gated: sub-8B models show modest
improvement and frequently degrade Marathi below the OCR
baseline.
\charbm{} outperforms domain-random selection by 2.8--4.0\,pp
absolute WER on Hindi across ${\geq}$8B models, matching or
exceeding dense retrieval on Hindi for 10 of 11 models with no
GPU and sub-millisecond query latency.
A persistent WER/CER tension reveals that word-level fluency
is recovered before character-exact restoration, most severely
in sub-8B models.

Several directions follow from this work.
Combining \charbm{} retrieval with inter-sentence context in
the prompt is the most immediate extension, given that
Bhandari and Harit~\cite{10.1145/3815575} showed large
gains from preceding-sentence context in the fine-tuned regime.
A hybrid retrieval strategy interpolating BM25 and dense
scores is motivated by the complementary behaviour of the two
methods across languages and model scales.
Extending evaluation to Gujarati, Bengali, Tamil, and Telugu
is a natural next step, as the \charbm{} principle applies
to any script where OCR errors manifest as predictable
character-level distortions.
Validation on real scanned documents and lightweight
fine-tuning (LoRA) with \charbm{}-retrieved examples remain
important open questions.

\section*{Funding}

The authors declare that no funds, grants, or other financial support
were received during the preparation of this manuscript.

\FloatBarrier 
\bibliographystyle{splncs04}
\bibliography{refs}

\end{document}